%% file: acl_latex.tex
\documentclass[11pt]{article}

\input{preamble}
\usepackage[final]{acl_latex}

\newcommand{\gr}{\cellcolor{gray!12}}
\usepackage{times}
\usepackage{latexsym}
\usepackage{pifont}
\usepackage[T1]{fontenc}
\definecolor{lightgrayblue}{RGB}{221,223,232}
\usepackage{makecell}
\usepackage[utf8]{inputenc}

\usepackage{microtype}

\usepackage{inconsolata}

\usepackage{graphicx}
\usepackage{titlesec}
\usepackage[most]{tcolorbox}
\newtcbox{\hlredtabA}{on line, box align=base, colback=red!10, colframe=white, size=fbox, arc=3pt, 
	before upper=\strut, top=-2pt, bottom=-4pt, left=-2pt, right=-2pt, boxrule=0pt}
\newtcbox{\hlredtabB}{on line, box align=base, colback=red!20, colframe=white, size=fbox, arc=3pt, 
	before upper=\strut, top=-2pt, bottom=-4pt, left=-2pt, right=-2pt, boxrule=0pt}
\newtcbox{\hlredtabC}{on line, box align=base, colback=red!30, colframe=white, size=fbox, arc=3pt, 
	before upper=\strut, top=-2pt, bottom=-4pt, left=-2pt, right=-2pt, boxrule=0pt}
\newtcbox{\hlredtabD}{on line, box align=base, colback=red!40, colframe=white, size=fbox, arc=3pt, 
	before upper=\strut, top=-2pt, bottom=-4pt, left=-2pt, right=-2pt, boxrule=0pt}
\newtcbox{\hlredtabE}{on line, box align=base, colback=red!50, colframe=white, size=fbox, arc=3pt, 
	before upper=\strut, top=-2pt, bottom=-4pt, left=-2pt, right=-2pt, boxrule=0pt}
\newtcbox{\hlredtabF}{on line, box align=base, colback=red!60, colframe=white, size=fbox, arc=3pt, 
	before upper=\strut, top=-2pt, bottom=-4pt, left=-2pt, right=-2pt, boxrule=0pt}
\newtcbox{\hlredtabG}{on line, box align=base, colback=red!70, colframe=white, size=fbox, arc=3pt, 
	before upper=\strut, top=-2pt, bottom=-4pt, left=-2pt, right=-2pt, boxrule=0pt}
\newtcbox{\hlredtabH}{on line, box align=base, colback=red!80, colframe=white, size=fbox, arc=3pt, 
	before upper=\strut, top=-2pt, bottom=-4pt, left=-2pt, right=-2pt, boxrule=0pt}
\newtcbox{\hlredtabI}{on line, box align=base, colback=red!90, colframe=white, size=fbox, arc=3pt, 
	before upper=\strut, top=-2pt, bottom=-4pt, left=-2pt, right=-2pt, boxrule=0pt}
\newtcbox{\hlredtabJ}{on line, box align=base, colback=red!100, colframe=white, size=fbox, arc=3pt, 
	before upper=\strut, top=-2pt, bottom=-4pt, left=-2pt, right=-2pt, boxrule=0pt}

\newtcbox{\hlgreentabA}{on line, box align=base, colback=green!5, colframe=white, size=fbox, arc=3pt, 
	before upper=\strut, top=-2pt, bottom=-4pt, left=-2pt, right=-2pt, boxrule=0pt}
\newtcbox{\hlgreentabB}{on line, box align=base, colback=green!10, colframe=white, size=fbox, arc=3pt, 
	before upper=\strut, top=-2pt, bottom=-4pt, left=-2pt, right=-2pt, boxrule=0pt}
\newtcbox{\hlgreentabC}{on line, box align=base, colback=green!15, colframe=white, size=fbox, arc=3pt, 
	before upper=\strut, top=-2pt, bottom=-4pt, left=-2pt, right=-2pt, boxrule=0pt}
\newtcbox{\hlgreentabD}{on line, box align=base, colback=green!20, colframe=white, size=fbox, arc=3pt, 
	before upper=\strut, top=-2pt, bottom=-4pt, left=-2pt, right=-2pt, boxrule=0pt}
\newtcbox{\hlgreentabE}{on line, box align=base, colback=green!25, colframe=white, size=fbox, arc=3pt, 
	before upper=\strut, top=-2pt, bottom=-4pt, left=-2pt, right=-2pt, boxrule=0pt}
\newtcbox{\hlgreentabF}{on line, box align=base, colback=green!30, colframe=white, size=fbox, arc=3pt, 
	before upper=\strut, top=-2pt, bottom=-4pt, left=-2pt, right=-2pt, boxrule=0pt}
\newtcbox{\hlgreentabG}{on line, box align=base, colback=green!35, colframe=white, size=fbox, arc=3pt, 
	before upper=\strut, top=-2pt, bottom=-4pt, left=-2pt, right=-2pt, boxrule=0pt}
\newtcbox{\hlgreentabH}{on line, box align=base, colback=green!40, colframe=white, size=fbox, arc=3pt, 
	before upper=\strut, top=-2pt, bottom=-4pt, left=-2pt, right=-2pt, boxrule=0pt}
\newtcbox{\hlgreentabI}{on line, box align=base, colback=green!45, colframe=white, size=fbox, arc=3pt, 
	before upper=\strut, top=-2pt, bottom=-4pt, left=-2pt, right=-2pt, boxrule=0pt}
\newtcbox{\hlgreentabJ}{on line, box align=base, colback=green!50, colframe=white, size=fbox, arc=3pt, 
	before upper=\strut, top=-2pt, bottom=-4pt, left=-2pt, right=-2pt, boxrule=0pt}

\newcommand{\dalgshifted}{\raisebox{0.5\depth}{$\downarrow$}}
\newcommand{\daugshifted}{\raisebox{0.5\depth}{$\uparrow$}}

\newcommand{\uaA}[1]{\hlgreentabA{\scriptsize \daugshifted{#1\%}}} % 0-10%
\newcommand{\uaB}[1]{\hlgreentabB{\scriptsize \daugshifted{#1\%}}} % 10-20%
\newcommand{\uaC}[1]{\hlgreentabC{\scriptsize \daugshifted{#1\%}}} % 20-30%
\newcommand{\uaD}[1]{\hlgreentabD{\scriptsize \daugshifted{#1\%}}} % 30-40%
\newcommand{\uaE}[1]{\hlgreentabE{\scriptsize \daugshifted{#1\%}}} % 40-50%
\newcommand{\uaF}[1]{\hlgreentabF{\scriptsize \daugshifted{#1\%}}} % 50-60%
\newcommand{\uaG}[1]{\hlgreentabG{\scriptsize \daugshifted{#1\%}}} % 60-70%
\newcommand{\uaH}[1]{\hlgreentabH{\scriptsize \daugshifted{#1\%}}} % 70-80%
\newcommand{\uaI}[1]{\hlgreentabI{\scriptsize \daugshifted{#1\%}}} % 80-90%
\newcommand{\uaJ}[1]{\hlgreentabJ{\scriptsize \daugshifted{#1\%}}} % 90-100%

\newcommand{\daA}[1]{\hlredtabA{\scriptsize \dalgshifted{#1\%}}} % 0-10%
\newcommand{\daB}[1]{\hlredtabB{\scriptsize \dalgshifted{#1\%}}} % 10-20%
\newcommand{\daC}[1]{\hlredtabC{\scriptsize \dalgshifted{#1\%}}} % 20-30%
\newcommand{\daD}[1]{\hlredtabD{\scriptsize \dalgshifted{#1\%}}} % 30-40%
\newcommand{\daE}[1]{\hlredtabE{\scriptsize \dalgshifted{#1\%}}} % 40-50%
\newcommand{\daF}[1]{\hlredtabF{\scriptsize \dalgshifted{#1\%}}} % 50-60%
\newcommand{\daG}[1]{\hlredtabG{\scriptsize \dalgshifted{#1\%}}} % 60-70%
\newcommand{\daH}[1]{\hlredtabH{\scriptsize \dalgshifted{#1\%}}} % 70-80%
\newcommand{\daI}[1]{\hlredtabI{\scriptsize \dalgshifted{#1\%}}} % 80-90%
\newcommand{\daJ}[1]{\hlredtabJ{\scriptsize \dalgshifted{#1\%}}} % 90-100%

\newcommand{\ColorValue}[1]{%
	\pgfmathparse{#1}%
	\ifdim\pgfmathresult pt>90pt
	\daJ{#1}%
	\else
	\ifdim\pgfmathresult pt>80pt
	\daI{#1}%
	\else
	\ifdim\pgfmathresult pt>70pt
	\daH{#1}%
	\else
	\ifdim\pgfmathresult pt>60pt
	\daG{#1}%
	\else
	\ifdim\pgfmathresult pt>50pt
	\daF{#1}%
	\else
	\ifdim\pgfmathresult pt>40pt
	\daE{#1}%
	\else
	\ifdim\pgfmathresult pt>30pt
	\daD{#1}%
	\else
	\ifdim\pgfmathresult pt>20pt
	\daC{#1}%
	\else
	\ifdim\pgfmathresult pt>10pt
	\daB{#1}%
	\else
	\ifdim\pgfmathresult pt>0pt
	\daA{#1}%
	\fi
	\fi
	\fi
	\fi
	\fi
	\fi
	\fi
	\fi
	\fi
	\fi
}

\newcommand{\ua}[1]{%
	\pgfmathparse{#1}%
	\ifdim\pgfmathresult pt>90pt
	\uaJ{#1}%
	\else
	\ifdim\pgfmathresult pt>80pt
	\uaI{#1}%
	\else
	\ifdim\pgfmathresult pt>70pt
	\uaH{#1}%
	\else
	\ifdim\pgfmathresult pt>60pt
	\uaG{#1}%
	\else
	\ifdim\pgfmathresult pt>50pt
	\uaF{#1}%
	\else
	\ifdim\pgfmathresult pt>40pt
	\uaE{#1}%
	\else
	\ifdim\pgfmathresult pt>30pt
	\uaD{#1}%
	\else
	\ifdim\pgfmathresult pt>20pt
	\uaC{#1}%
	\else
	\ifdim\pgfmathresult pt>10pt
	\uaB{#1}%
	\else
	\ifdim\pgfmathresult pt>0pt
	\uaA{#1}%
	\fi
	\fi
	\fi
	\fi
	\fi
	\fi
	\fi
	\fi
	\fi
	\fi
}

\newtcbox{\hlyellowtabA}{on line, box align=base, colback=gray!5, colframe=white, size=fbox, arc=3pt, before upper=\strut, top=-2pt, bottom=-4pt, left=-2pt, right=-2pt, boxrule=0pt}
\newtcbox{\hlyellowtabB}{on line, box align=base, colback=gray!10, colframe=white, size=fbox, arc=3pt, before upper=\strut, top=-2pt, bottom=-4pt, left=-2pt, right=-2pt, boxrule=0pt}
\newtcbox{\hlyellowtabC}{on line, box align=base, colback=gray!15, colframe=white, size=fbox, arc=3pt, before upper=\strut, top=-2pt, bottom=-4pt, left=-2pt, right=-2pt, boxrule=0pt}
\newtcbox{\hlyellowtabD}{on line, box align=base, colback=gray!20, colframe=white, size=fbox, arc=3pt, before upper=\strut, top=-2pt, bottom=-4pt, left=-2pt, right=-2pt, boxrule=0pt}
\newtcbox{\hlyellowtabE}{on line, box align=base, colback=gray!25, colframe=white, size=fbox, arc=3pt, before upper=\strut, top=-2pt, bottom=-4pt, left=-2pt, right=-2pt, boxrule=0pt}
\newtcbox{\hlyellowtabF}{on line, box align=base, colback=gray!30, colframe=white, size=fbox, arc=3pt, before upper=\strut, top=-2pt, bottom=-4pt, left=-2pt, right=-2pt, boxrule=0pt}
\newtcbox{\hlyellowtabG}{on line, box align=base, colback=gray!35, colframe=white, size=fbox, arc=3pt, before upper=\strut, top=-2pt, bottom=-4pt, left=-2pt, right=-2pt, boxrule=0pt}
\newtcbox{\hlyellowtabH}{on line, box align=base, colback=gray!40, colframe=white, size=fbox, arc=3pt, before upper=\strut, top=-2pt, bottom=-4pt, left=-2pt, right=-2pt, boxrule=0pt}
\newtcbox{\hlyellowtabI}{on line, box align=base, colback=gray!45, colframe=white, size=fbox, arc=3pt, before upper=\strut, top=-2pt, bottom=-4pt, left=-2pt, right=-2pt, boxrule=0pt}
\newtcbox{\hlyellowtabJ}{on line, box align=base, colback=gray!50, colframe=white, size=fbox, arc=3pt, before upper=\strut, top=-2pt, bottom=-4pt, left=-2pt, right=-2pt, boxrule=0pt}

\newcommand{\deltayellow}[1]{%
	\pgfmathparse{#1}%
	\ifdim\pgfmathresult pt>90pt \hlyellowtabJ{\scriptsize $\Delta$#1}
	\else\ifdim\pgfmathresult pt>80pt \hlyellowtabI{\scriptsize $\Delta$#1}
	\else\ifdim\pgfmathresult pt>70pt \hlyellowtabH{\scriptsize $\Delta$#1}
	\else\ifdim\pgfmathresult pt>60pt \hlyellowtabG{\scriptsize $\Delta$#1}
	\else\ifdim\pgfmathresult pt>50pt \hlyellowtabF{\scriptsize $\Delta$#1}
	\else\ifdim\pgfmathresult pt>40pt \hlyellowtabE{\scriptsize $\Delta$#1}
	\else\ifdim\pgfmathresult pt>30pt \hlyellowtabD{\scriptsize $\Delta$#1}
	\else\ifdim\pgfmathresult pt>20pt \hlyellowtabC{\scriptsize $\Delta$#1}
	\else\ifdim\pgfmathresult pt>10pt \hlyellowtabB{\scriptsize $\Delta$#1}
	\else\ifdim\pgfmathresult pt>0pt \hlyellowtabA{\scriptsize $\Delta$#1}
	\fi\fi\fi\fi\fi\fi\fi\fi\fi\fi
}

\title{Are We Using the Right Benchmark: \\An Evaluation Framework for  Visual Token Compression Methods}

\author{Chenfei Liao\textsuperscript{1,2,6} \quad Wensong Wang\textsuperscript{3,2} \quad Zichen Wen\textsuperscript{2,5} \quad Xu Zheng\textsuperscript{1,4,6}\\ \quad \textbf{Yiyu Wang\textsuperscript{2} 
\quad Haocong He\textsuperscript{2}  \quad Yuanhuiyi Lyu\textsuperscript{1,6} \quad Lutao Jiang\textsuperscript{1,6}  \quad Xin Zou\textsuperscript{1,6}}\\  \quad \textbf{Yuqian Fu\textsuperscript{4} \quad Bin Ren\textsuperscript{7}}  \quad 
\textbf{Linfeng Zhang\textsuperscript{2,}\footnotemark[2] \quad Xuming Hu\textsuperscript{1,6,}\footnotemark[2]}\\
\textsuperscript{1}The Hong Kong University of Science and Technology (Guangzhou) \\ \quad \textsuperscript{2}Shanghai Jiao Tong University   \quad \textsuperscript{3}Northeastern University\\   \quad \textsuperscript{4}INSAIT, Sofia University “St. Kliment Ohridski”  \quad \textsuperscript{5}Shanghai AI Laboratory \\ \quad \textsuperscript{6}The Hong Kong University of Science and Technology   \quad \textsuperscript{7}MBZUAI
}

\begin{document}

\maketitle

{
\renewcommand{\thefootnote}{\fnsymbol{footnote}}
\footnotetext[2]{Corresponding authors.}
}

\begin{figure*}[!h]
    \centering

    \includegraphics[width=0.95\linewidth]{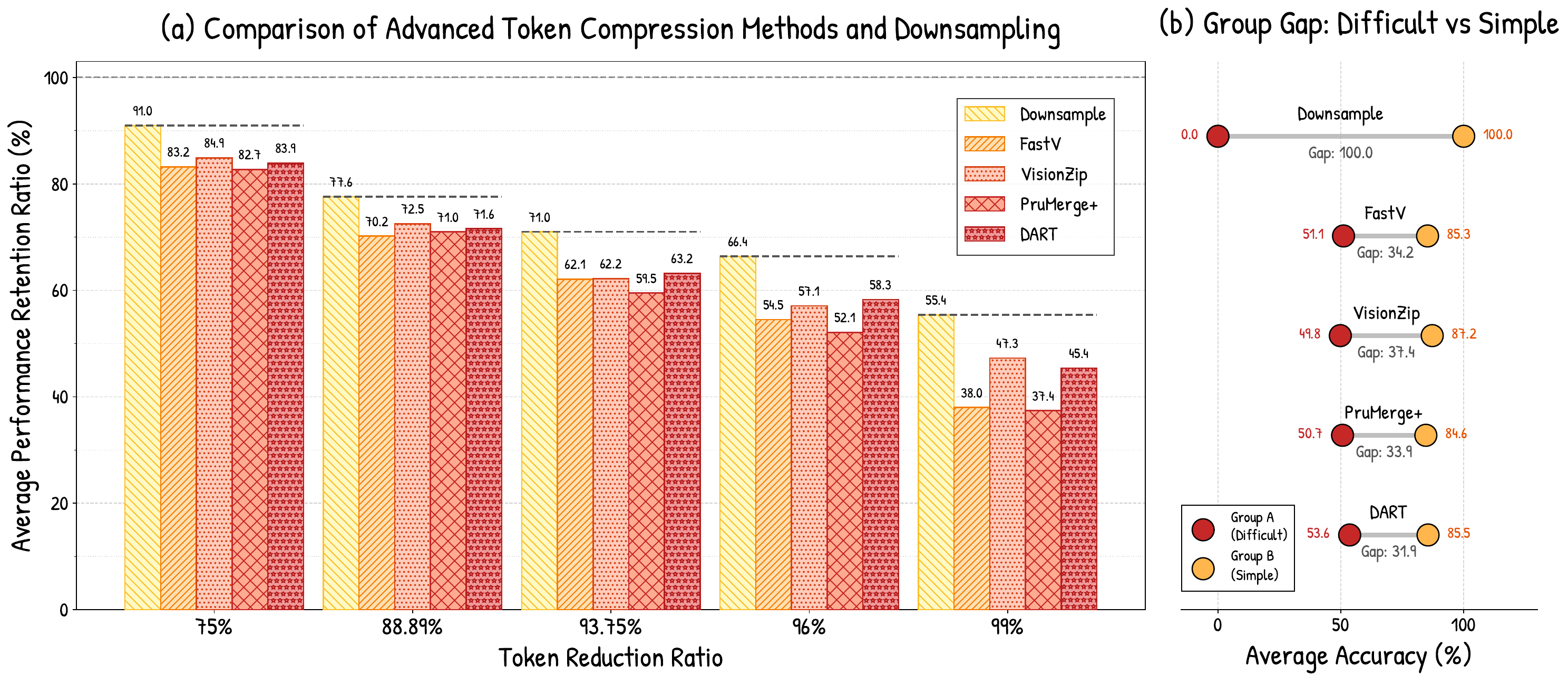}
    \caption{(a) Average Performance Retention Ratio (APRR) of five visual token compression methods on eight benchmarks (Model: Qwen2-VL-7B; Benchmark: as shown in Table \ref{table1}), supporting \textit{\textbf{the observation (i):  current benchmarks contain substantial noise (task-irrelevant samples) for evaluating visual token compression}}. (b) Comparison of advanced token compression methods and downsampling on Qwen2-VL-7B by groups at 75\% compression ratio, supporting \textit{\textbf{the observation (ii):   downsampling can act as an effective data filter that distinguishes between simple and difficult samples with respect to compression sensitivity}}. Group A refers to the ``difficult'' samples. Group B refers to the ``simple'' samples. The grouping strategy is explained in Section~\ref{hypothesis}.} 
    \label{fig:placeholder}
\end{figure*}

\input{sections/0_abstract}

\input{sections/1_intro}
\input{sections/2_related_work}

\input{sections/3_exp}

\input{sections/4_bench}
\input{sections/5_conclusion}

\bibliography{custom}

\input{sections/Append}

\end{document}

%% file: preamble.tex
\usepackage{amsmath,amsfonts}
\usepackage{algorithmic}
\usepackage{array}
\usepackage[caption=false,font=normalsize,labelfont=sf,textfont=sf]{subfig}
\usepackage{textcomp}
\usepackage{stfloats}
\usepackage{url}
\usepackage{verbatim}
\usepackage{graphicx}
\usepackage{tabularx}
\usepackage{pifont}
\usepackage{rotating}
\usepackage{hyperref}
\usepackage{xurl}  % 允许URL任意位置换行
\usepackage{hyperref}
\usepackage {wrapfig}
\usepackage{tikz}
\usepackage{multirow} 
\usepackage{booktabs}
\usepackage{adjustbox}
\usepackage{xcolor}
\usepackage{pifont}

\definecolor{lightgreen}{rgb}{0.56, 0.93, 0.56}
\definecolor{brightlavender}{rgb}{0.75, 0.58, 0.89}
\definecolor{capri}{rgb}{0.0, 0.75, 1.0}
\definecolor{darkpastelgreen}{rgb}{0.01, 0.75, 0.24}

\definecolor{tree-level-1}{RGB}{245,20,85}
\definecolor{tree-level-2}{RGB}{246,86,118}
\definecolor{tree-level-3}{RGB}{248,177,193}
\definecolor{tree-leaf}{RGB}{176,230,198}
\usepackage{dsfont}

\definecolor{titlecolor}{RGB}{0,0,0}
\definecolor{hidden-draw}{RGB}{0,0,0}
\definecolor{lighttealblue}{RGB}{0,0,0}    
\definecolor{lightplum}{RGB}{0, 0, 0}        
\definecolor{harvestgold}{RGB}{0, 0, 0}   

\usepackage{booktabs} % 用于 \toprule, \midrule, \bottomrule
\usepackage{colortbl} % 用于 \rowcolor

\definecolor{my_green}{RGB}{51,102,0}
\definecolor{my_red}{RGB}{204, 0, 0}
\usepackage[utf8]{inputenc} % allow utf-8 input
\usepackage[T1]{fontenc}    % use 8-bit T1 fonts
\usepackage{hyperref}       % hyperlinks
\usepackage{url}            % simple URL typesetting
\usepackage{booktabs}       % professional-quality tables
\usepackage{amsfonts}       % blackboard math symbols
\usepackage{nicefrac}       % compact symbols for 1/2, etc.
\usepackage{microtype}      % microtypography
\usepackage{xcolor}         % colors

%% file: sections/0_abstract.tex
\begin{abstract}
Recent efforts to accelerate inference in Multimodal Large Language Models (MLLMs) have largely focused on visual token compression. 
The effectiveness of these methods is commonly evaluated by measuring the accuracy drop on existing MLLM benchmarks before and after compression. However, these benchmarks are originally designed to assess general perception and reasoning abilities, rather than the specific challenges posed by visual token compression, leading to a fundamental task mismatch. In this work, we uncover a counterintuitive yet consistent phenomenon: \textbf{simple image downsampling outperforms many advanced visual token compression methods across multiple widely used benchmarks}. 
Through a comprehensive empirical study spanning eight popular benchmarks and multiple state-of-the-art compression techniques, we show that (i) current benchmarks contain substantial noise (task-irrelevant samples) for evaluating visual token compression, and (ii) downsampling can act as an effective data filter that distinguishes between simple and difficult samples with respect to compression sensitivity. Motivated by these findings, we propose \textbf{VTC-Bench}\footnote{Code: \href{https://github.com/Chenfei-Liao/VTC-Bench}{https://github.com/Chenfei-Liao/VTC-Bench}.}\footnote{Project Page: \href{https://chenfei-liao.github.io/VTC-Bench-Page}{https://chenfei-liao.github.io/VTC-Bench-Page}.}, an evaluation framework that explicitly leverages downsampling as a discriminator to denoise existing benchmarks, enabling a fairer and more meaningful additional assessment of visual token compression methods.% additional
%We hope this work provides clearer evaluation principles and fosters more reliable progress in efficient MLLM inference. 
%The code will be released to facilitate future research.
\end{abstract}

%% file: sections/1_intro.tex
\section{Introduction}

% Multi-modal Large Language Models (MLLMs) have demonstrated remarkable capabilities in understanding, reasoning, and generating content across vision and language modalities, enabling a wide range of applications such as embodied AI~\cite{yin2024survey,cheng2025embodiedeval}.
% However, the computational cost of processing images, especially those with high resolution, remains a significant bottleneck in employing and scaling these systems efficiently~\cite{liu2025shifting}. 
% This challenge arises from the fact that the visual tokens, which represent image patches, often vastly outnumber the textual tokens, resulting in substantial increases in memory usage and inference latency~\cite{wang2025effivlm}.
% To address this, numerous visual token compression methods have been proposed, aiming to reduce redundancy while preserving critical visual information~\cite{yang2025visionzip,xing2024pyramiddrop}.
% However, many of these methods are evaluated on general MLLM benchmarks~\cite{li2024survey}. 
% These benchmarks are not specifically designed and may not qualify as a suitable benchmark for visual token compression.
Multimodal Large Language Models (MLLMs) have shown impressive abilities in understanding, reasoning, and generating content across vision and language~\cite{chen2024mj,kang2025legion,zou2025look,chai2024auroracap,wen2026innovator}, enabling applications such as embodied AI~\cite{yin2024survey,fu2025objectrelator,zhang2025a4,zhang2025phystoolbench,zhang2026panoramic}.
However, their efficiency is often constrained by the high computational cost of processing images, particularly at high resolutions~\cite{liu2025shifting}. 
This bottleneck arises because visual tokens, derived from image patches, typically far outnumber textual tokens, leading to substantial memory consumption and inference latency~\cite{wang2025effivlm,chen2025variation,wen2025efficient}.
To mitigate this issue, numerous visual token compression methods have been proposed to reduce redundancy while retaining essential visual information~\cite{yang2025visionzip,xing2024pyramiddrop,xiong2025prune2drive,zou2025holov}. 

Yet, these methods are typically evaluated on general MLLM benchmarks~\cite{li2024survey}, which are not designed for compression, therefore failing to provide appropriate evaluation criteria.
Thus, in this paper, we uncover a surprising finding: as in Figure~\ref {fig:placeholder}, \textit{\textbf{simple image downsampling consistently outperforms many advanced compression methods across multiple widely used benchmarks.}}
This suggests that current evaluation frameworks don't adequately capture the challenges inherent in visual token compression. 

%In particular, we hypothesize that many existing benchmarks contain a substantial proportion of overly simplistic samples that do not require fine-grained visual understanding, thereby favoring methods that preserve global information at the expense of local detail.
To investigate this, we conduct a comprehensive empirical study comparing multiple state-of-the-art visual token compression methods against a simple downsampling baseline across eight widely used benchmarks. 
Based on the study results in Table \ref{table1} and \ref{table2}, two crucial findings are concluded:
\ding{172} The counterintuitive phenomenon mentioned above generally exists in popular benchmarks, proving that \textit{\textbf{current benchmarks are noisy for the visual token compression task}}. Many task-irrelevant samples serve as noise when evaluating visual token compression methods.
\ding{173} The correct sample group under downsampling methods has achieved significantly better accuracy than the incorrect sample group under downsampling methods across various compression methods and benchmarks, proving that \textit{\textbf{downsampling can serve as a data filter to evaluate the difficulty of samples upon the visual token compression task.}}
%Our results confirm that downsampling consistently achieves superior performance on general-purpose tasks (e.g., MMBench, GQA), while specialized methods excel only in tasks requiring precise visual details (e.g., ChartQA, DocVQA). %This divergence highlights a systematic bias in benchmark composition, which may mislead the development and evaluation of future compression algorithms.

Based on these findings, we propose VTC-Bench, an evaluation framework specifically designed to optimize current existing benchmarks, aiming to serve as an additional reference benchmark for evaluating visual token compression methods.
By explicitly distinguishing between “simple” and “difficult” samples through downsampling, VTC-Bench adaptively and fairly selects ``difficult'' samples that satisfy the requirements of evaluating visual token compression methods.

%[experiments on evaluating the effectiveness of our framework]
Overall, our contributions are threefold:
\ding{172} We identify and validate the data noise (task-irrelevant samples) in existing MLLM benchmarks on the visual token compression task.
\ding{173} We introduce a data filtering mechanism using downsampling as a discriminator to categorize benchmark samples by difficulty.
\ding{174} We propose VTC-Bench, the first evaluation framework tailored for fairly evaluating visual token compression methods, aiming to foster more meaningful progress in this emerging field.

%% file: sections/2_related_work.tex
\begin{table*}[h!]
\small
\renewcommand{\arraystretch}{1.1}
\centering
\caption{Comparison of Advanced Token Compression Methods and Downsampling on Qwen2-VL-7B. APRR refers to  \textbf{\textit{the average performance retention ratio}}, which is the average value of the retention ratio compared to the vanilla performance, across the eight benchmarks. The right side of each cell shows \textbf{\textit{the relative change percentage compared to the performance of the downsampling method}}. }
\label{table1}
\renewcommand{\tabcolsep}{8pt}
\resizebox{\textwidth}{!}{
\begin{tabular}{l|c c c c c c c c |c}
\toprule
\textit{\textbf{Method}} & \textit{\textbf{GQA} }& \textit{\textbf{MMB}} & \textit{\textbf{MMB$^{CN}$} }& \textit{\textbf{MME}} & \textit{\textbf{POPE}} & \textit{\textbf{MMStar}}  & \textit{\textbf{OCRBench}}& \textit{\textbf{ChartQA}}  & \textit{\textbf{APRR}}\\
\midrule
\rowcolor{lightgrayblue} \textit{Qwen2-VL-7B~\cite{wang2024qwen2}} & \multicolumn{9}{c}{\textit{Upper Bound. All Tokens (100\%)}} \\ \midrule
\textcolor{gray}{Vanilla} & \textcolor{gray}{62.3} & \textcolor{gray}{78.9} & \textcolor{gray}{78.0} & \textcolor{gray}{2306} & \textcolor{gray}{88.4} & \textcolor{gray}{57.1} & \textcolor{gray}{80.7}&  \textcolor{gray}{81.6} & \textcolor{gray}{100.0}\\
\midrule
\rowcolor{lightgrayblue}\textit{Qwen2-VL-7B~\cite{wang2024qwen2}} & \multicolumn{9}{c}{\textit{Token Reduction ($\downarrow$ 75.00\%)}} \\ \midrule
\rowcolor{brown!10}
+ Downsample (Baseline) & \underline{59.2}   & \textbf{75.0} & \textbf{73.8} & \textbf{2259}&  86.2& \textbf{50.1} &\textbf{64.9}  &\underline{65.0} & \textbf{91.0}   \\
+ FastV~\cite{chen2024image} & 57.0 \ColorValue{3.7}& \underline{73.7} \ColorValue{1.7} & \underline{73.1} \ColorValue{0.9}& \underline{2083} \ColorValue{7.8}& 84.5 \ColorValue{2.0}&44.6 \ColorValue{11.0}&42.0 \ColorValue{35.3} &58.1 \ColorValue{10.6}& 83.2 \ColorValue{8.6} \\
+ VisionZip~\cite{yang2025visionzip} &  58.6 \ColorValue{1.0}& 71.1 \ColorValue{5.2}& 70.5 \ColorValue{4.5} & 2062 \ColorValue{8.7}& \underline{87.1} \ua{1.0} & 47.2 \ColorValue{5.8} &42.1 \ColorValue{35.1}&\textbf{66.9} \ua{2.9} & \underline{84.9} \ColorValue{6.7}\\
+ PruMerge+~\cite{shang2024llava} &  \textbf{59.4} \ua{0.3}&  72.1 \ColorValue{3.9}& 72.0 \ColorValue{2.4} & 2044 \ColorValue{9.5} &\textbf{87.2} \ua{1.2}  &  \underline{48.0} \ColorValue{4.2}&33.9 \ColorValue{47.8}&56.2 \ColorValue{13.5}& 82.7 \ColorValue{9.1}\\
+ DART~\cite{wen2025stop}   & 56.9 \ColorValue{3.9} &  72.5 \ColorValue{3.3}&  70.2 \ColorValue{4.9}& 2066 \ColorValue{8.5}& 84.7  \ColorValue{1.7} & 47.2 \ColorValue{5.8}&\underline{52.5} \ColorValue{19.1}&52.7 \ColorValue{18.9} & 83.9 \ColorValue{7.8}  \\
\midrule
\rowcolor{lightgrayblue} \textit{Qwen2-VL-7B~\cite{wang2024qwen2}} & \multicolumn{9}{c}{\textit{Token Reduction ($\downarrow$ 88.89\%)}} \\ \midrule
\rowcolor{brown!10}
+ Downsample (Baseline)  &  \textbf{55.5}& \textbf{69.0} & \textbf{70.2} &\textbf{2127} &82.9  & \textbf{44.0} & \textbf{48.8} & 24.8 & \textbf{77.6}  \\
+ FastV~\cite{chen2024image}  &52.3 \ColorValue{5.8} & 65.0 \ColorValue{5.8} & \underline{65.5} \ColorValue{6.7} & 1854 \ColorValue{12.8} &  77.4 \ColorValue{6.6} & \underline{40.3} \ColorValue{8.4} & 25.9 \ColorValue{46.9} & 32.9 \ua{32.6} & 70.2 \ColorValue{9.5} \\
+ VisionZip~\cite{yang2025visionzip} & 53.3 \ColorValue{4.0} & 62.9 \ColorValue{8.8} & 63.0 \ColorValue{10.3} & 1820 \ColorValue{14.4} & \underline{83.6} \ua{0.8} & 40.2 \ColorValue{8.6}  & 25.1 \ColorValue{48.6} & \textbf{48.4} \ua{95.2} & \underline{72.5} \ColorValue{6.6}\\
+ PruMerge+~\cite{shang2024llava} & \underline{54.8} \ColorValue{1.3} & 62.2 \ColorValue{9.9} & 61.3 \ColorValue{12.7} & 1806 \ColorValue{15.1} & \textbf{84.3} \ua{1.7} & 38.4 \ColorValue{12.7} & 22.2 \ColorValue{54.5} & \underline{44.2} \ua{78.2} & 71.0 \ColorValue{8.5} \\
+ DART~\cite{wen2025stop} & 51.9 \ColorValue{6.5} & 61.3 \ColorValue{11.2} & 61.8 \ColorValue{12.0} & \underline{1915} \ColorValue{10.0} & 80.5 \ColorValue{2.9} & 39.8 \ColorValue{9.5} &\underline{41.0} \ColorValue{16.0} & 30.8 \ua{24.2} & 71.6 \ColorValue{7.7} \\
\midrule
\rowcolor{lightgrayblue} \textit{Qwen2-VL-7B~\cite{wang2024qwen2}} & \multicolumn{9}{c}{\textit{Token Reduction ($\downarrow$ 93.75\%)}} \\ \midrule
\rowcolor{brown!10}
+ Downsample (Baseline)   &\textbf{52.6}  & \textbf{66.4} & \textbf{66.8} &\textbf{1994} &\underline{79.5} & \textbf{40.9} &\textbf{40.3}&12.7 & \textbf{71.0}\\
+ FastV~\cite{chen2024image}  & 49.0 \ColorValue{6.8} & \underline{57.1} \ColorValue{14.0} & \underline{57.9} \ColorValue{13.3} & 1684 \ColorValue{15.5} & 74.9 \ColorValue{5.8} & \underline{37.5} \ColorValue{8.3} & 18.7 \ColorValue{53.6} & 20.6 \ua{62.2} & 62.1 \ColorValue{12.5}\\
+ VisionZip~\cite{yang2025visionzip} & 49.0 \ColorValue{6.8} & 54.8 \ColorValue{17.5} &54.0 \ColorValue{19.2} & 1704 \ColorValue{14.5} & \textbf{80.2} \ColorValue{0.9} &35.2 \ColorValue{13.9} &15.9 \ColorValue{60.5} &\underline{28.0} \ua{120.5} &62.2 \ColorValue{12.4}\\
+ PruMerge+~\cite{shang2024llava} &  48.7 \ColorValue{7.4} &  48.4 \ColorValue{27.1} & 48.1 \ColorValue{28.0} &1679 \ColorValue{15.8} & 79.2 \ColorValue{0.4} & 33.2 \ColorValue{18.8} &14.4 \ColorValue{64.3} &\textbf{30.0} \ua{136.2} & 59.5 \ColorValue{16.2}\\
+ DART~\cite{wen2025stop}  &\underline{49.2} \ColorValue{6.5} & 53.4 \ColorValue{19.6} & 54.0 \ColorValue{19.2} &\underline{1786} \ColorValue{10.4} & 78.1 \ColorValue{1.8} & 33.6 \ColorValue{17.8} & \underline{33.7} \ColorValue{16.4} & 19.2 \ua{51.2} & \underline{63.2} \ColorValue{11.0}\\
\midrule
\rowcolor{lightgrayblue} \textit{Qwen2-VL-7B~\cite{wang2024qwen2}}& \multicolumn{9}{c}{\textit{Token Reduction ($\downarrow$ 96.00\%)}} \\ \midrule
\rowcolor{brown!10}
+ Downsample (Baseline)  & \textbf{50.1} &\textbf{62.0}  &\textbf{61.4}  & \textbf{1938}&  \textbf{78.8}& \textbf{37.5} &\textbf{32.3} &11.7 &\textbf{66.4}\\
+ FastV~\cite{chen2024image}  & 46.1 \ColorValue{8.0} & 43.9 \ColorValue{29.2} & 46.6 \ColorValue{24.1} & 1589 \ColorValue{18.0} & 72.4 \ColorValue{8.1} & \underline{33.6} \ColorValue{10.4} & 14.4 \ColorValue{55.4} &15.8 \ua{35.0} & 54.5 \ColorValue{17.9} \\
+ VisionZip~\cite{yang2025visionzip} & \underline{46.4} \ColorValue{7.4} &  \underline{49.5} \ColorValue{20.2} & \underline{50.0} \ColorValue{18.6} & 1628 \ColorValue{16.0} & \underline{77.8} \ColorValue{1.3} &33.4 \ColorValue{10.9} & 12.0 \ColorValue{62.8} &  \underline{19.4} \ua{65.8} & 57.1 \ColorValue{14.0} \\
+ PruMerge+~\cite{shang2024llava} &45.0 \ColorValue{10.2} &  39.1 \ColorValue{36.9} &40.9 \ColorValue{33.4} &1544 \ColorValue{20.3} &  74.0 \ColorValue{6.1} &30.5 \ColorValue{18.7} &10.5 \ColorValue{67.5} &  \textbf{20.9} \ua{78.6} & 52.1 \ColorValue{21.5} \\
+ DART~\cite{wen2025stop}  &  45.6 \ColorValue{9.0} & 47.9 \ColorValue{22.7} & 48.2 \ColorValue{21.5} & \underline{1701} \ColorValue{12.2} & 74.7 \ColorValue{5.2} & 31.7 \ColorValue{15.5} & \underline{29.3} \ColorValue{9.3} &16.6 \ua{41.9} & \underline{58.3} \ColorValue{12.2}\\
\midrule
\rowcolor{lightgrayblue} \textit{Qwen2-VL-7B~\cite{wang2024qwen2}} & \multicolumn{9}{c}{\textit{Token Reduction ($\downarrow$ 99.00\%)}} \\ \midrule
\rowcolor{brown!10}
+ Downsample (Baseline)  & \textbf{43.5} &  \textbf{51.6}& \textbf{51.9} &\textbf{1589} & \textbf{72.8} & \textbf{33.8}  & \underline{13.2}& 12.1 & \textbf{55.4} \\
+ FastV~\cite{chen2024image}  &38.2 \ColorValue{12.2} & 23.9  \ColorValue{53.7} & 24.5 \ColorValue{52.8} & 1189 \ColorValue{25.2} &  55.0 \ColorValue{24.5} & 26.1 \ColorValue{22.8} & 5.8 \ColorValue{56.1} & 11.9 \ColorValue{1.7} & 38.0 \ColorValue{31.4} \\
+ VisionZip~\cite{yang2025visionzip} & \underline{41.9} \ColorValue{3.7} & \underline{40.5} \ColorValue{21.5} & \underline{40.5} \ColorValue{22.0} & 1335 \ColorValue{16.0} & \underline{65.5} \ColorValue{10.0} & \underline{30.8} \ColorValue{8.9} & 4.9 \ColorValue{62.9} &\underline{12.8} \ua{5.9} &\underline{47.3} \ColorValue{14.6} \\
+ PruMerge+~\cite{shang2024llava} & 39.0 \ColorValue{10.3} & 23.7 \ColorValue{54.1} & 24.4 \ColorValue{53.0} &1165 \ColorValue{26.7} & 51.6 \ColorValue{29.1} & 25.7 \ColorValue{24.0} &3.5 \ColorValue{73.5} & \textbf{13.9} \ua{14.9} & 37.4 \ColorValue{32.5} \\
+ DART~\cite{wen2025stop}  & 40.5 \ColorValue{6.9} & 30.8 \ColorValue{40.3} & 30.7 \ColorValue{40.8} &\underline{1346} \ColorValue{15.3} & 60.0 \ColorValue{17.6} &28.8 \ColorValue{14.8} & \textbf{23.2} \ua{75.8} & 11.8 \ColorValue{2.5} & 45.4 \ColorValue{18.1}\\
\bottomrule
\end{tabular}}
\end{table*}
\section{Related Work}
\subsection{Visual Token Compression for MLLMs}
\label{ToCo}
Since visual tokens typically outnumber text tokens in MLLMs, compressing visual tokens has emerged as a promising strategy to accelerate inference~\cite{yao2026towards,shao2025survey,wu2026data,liu2025shifting, yang2025visionzip, wang2025effivlm, shang2024llava}. Leveraging the inherent redundancy in visual tokens, a variety of \textbf{\textit{training-free}} methods have been proposed.  
FastV~\cite{chen2024image}, the first to explore visual token compression in MLLMs, prunes redundant tokens based on their average attention scores. 
%Building on this idea, SparseVLM~\cite{zhangsparsevlm} introduces a recycling strategy to achieve more compact and flexible compression. 
Other subsequent methods mainly have two technical features: a) dividing the compression process into multiple stages~\cite{xing2024pyramiddrop, han2024filter, liu2024multi, chen2024efficient, endo2025feather,chen2025ipcv} to enable more precise identification of redundant tokens; b) seeking better metrics as the basis for visual token compression~\cite{han2024filter,zhangsparsevlm, xu2024freepruner, zou2025holov, wen2025stop,zheng2025towards}.
Specifically, VFlowOpt~\cite{yang2025vflowopt} introduces the importance map and a recycling mechanism to achieve a progressive and effective token pruning.
Similarly, G-Prune~\cite{jiang2025kind} identifies critical tokens through a graph-based perspective, while LUVC~\cite{zheng2025towards} applies a spectrum pruning unit to LLM, gradually pruning redundant visual tokens. 
In contrast, DART~\cite{wen2025stop} instead prioritizes token duplication as a key criterion, achieving surprisingly strong compression performance. 
Beyond these, GreedyPrune~\cite{pei2025greedyprune} and ToDRE~\cite{li2025todre} cast token compression as an optimization problem and employ greedy algorithms to search for efficient pruning strategies.

However, as in Section~\ref{Exp}, we have a surprising observation: across most MLLM benchmarks, these sophisticated visual token compression methods underperform compared to simply reducing the original image resolution, which motivates a deeper investigation into the underlying causes.

\subsection{MLLM Benchmarks}
% Benchmarks serve as a fundamental and indispensable support in the rapid development process of MLLMs. Current benchmarks mostly lay emphasis on fields like perception, reasoning, and so on~\cite{li2024survey}. For example, benchmarks like MME~\cite{yin2024survey}, MMBench~\cite{liu2024mmbench}, SEED-Bench~\cite{li2024seed}, MM-Vet~\cite{yu2023mm,yu2024mm}, and so on, offer comprehensive perception benchmarks, bringing tools to evaluate MLLMs' general ability to recognize visual information. Meanwhile, several domain-specific benchmarks are proposed to promote MLLMs' applications in scenarios like autonomous driving~\cite{sima2024drivelm,qian2024nuscenes}, remote sensing~\cite{muhtar2024lhrs}, and so on. When it comes to the token compression task in MLLMs, there is only one related benchmark~\cite{wang2025effivlm}, namely EffiVLM. EffiVLM offers a unified evaluation framework to benchmark training-free acceleration methods for LVLMs. EffiVLM evaluates acceleration methods based on the existing datasets, such as DocVQA~\cite{mathew2021docvqa} and ChartQA~\cite{masry2022chartqa}, with no targeted adjustments in data based on task-specific characteristics. In our paper, we attempt to explore some data-related rules in MLLM token compression tasks. Based on these rules, we propose VTC-Bench, the first challenging benchmark targeted at visual token compression for MLLMs, hoping to bring new insights to this field.
% Benchmarks play a fundamental role in the era of MLLMs while e
Existing MLLM benchmarks primarily focus on areas such as perception and reasoning~\cite{li2024survey,song2025video}. For example, MME~\cite{yin2024survey}, MMBench~\cite{liu2024mmbench}, and MM-Vet~\cite{yu2023mm,yu2024mm} provide broad perception-focused evaluations of MLLMs' visual understanding. In parallel, domain benchmarks target specific applications such as autonomous driving~\cite{sima2024drivelm,qian2024nuscenes} and remote sensing~\cite{muhtar2024lhrs}.
For visual token compression in MLLMs, only one benchmark currently exists: EffiVLM~\cite{wang2025effivlm}. It offers a unified framework for benchmarking training-free acceleration methods but relies on existing datasets (e.g., DocVQA~\cite{mathew2021docvqa}, ChartQA~\cite{masry2022chartqa}) rather than data tailored to token compression. Building on data-driven insights into compression behavior, we introduce \textit{\textbf{VTC-Bench}}, the first dedicated, challenging evaluation framework as an additional tool for evaluating visual token compression in MLLMs.
We aim for VTC-Bench to catalyze new research and insights, enabling fair comparisons and sharper evaluations of token-compression methods.

%% file: sections/3_exp.tex
\section{Experiments \& Findings}
\label{Exp}

\subsection{Motivation}
\label{motivation}
Some recent MLLMs, such as Qwen2-VL~\cite{wang2024qwen2} and Qwen2.5-VL~\cite{bai2025qwen2}, natively support inputs of varying resolutions.  
A trivial yet efficient method to handle high-resolution images is to simply downsample them to a lower resolution, effectively using naive pixel sampling as a form of compression. 
However, as shown in Section~\ref {ToCo}, most token compression methods for MLLMs choose to adaptively drop useless tokens or merge similar tokens instead of directly downsampling the original image, which should be more intelligent and reasonable methods. 
While in recent works~\cite{yang2025visionthink}, it is surprising that \textbf{\textit{image downsampling exceeds other sophisticated methods under some settings.}} 
VisionThink~\cite{yang2025visionzip} considers this finding as a motivation to design more efficient MLLMs. 
Different from VisionThink~\cite{yang2025visionzip}, we would like to further investigate the causes of this anomalous phenomenon, thus deciding to comprehensively compare the results of the downsampling methods with other methods under various settings.

\begin{table*}[h]
\small
\renewcommand{\arraystretch}{1.3}
\centering
\caption{Comparison of advanced token compression methods on Qwen2-VL-7B. Values are formatted as: \textbf{Group A} ({\scriptsize Group B}) with difference $\Delta$ below. $\Delta$ refers to \textbf{\textit{the absolute gap between groups.}}}
\label{table2}
\renewcommand{\tabcolsep}{5pt}
\resizebox{\textwidth}{!}{
\begin{tabular}{l|c c c c c c c c |c}
\toprule
\textit{\textbf{Method}} & \textit{\textbf{GQA}} & \textit{\textbf{MMB}} & \textit{\textbf{MMB$^{CN}$}} & \textit{\textbf{MME}} & \textit{\textbf{POPE}} & \textit{\textbf{MMStar}}  & \textit{\textbf{OCRBench}}& \textit{\textbf{ChartQA}} & \textit{\textbf{Average}} \\
\midrule

% --- Section 1: 75.00% ---
\rowcolor{lightgrayblue}  \textit{\textbf{Group A} {\scriptsize (Group B)}} & \multicolumn{9}{c}{\textit{Token Reduction ($\downarrow$ 75.00\%)}} \\ \midrule
\rowcolor{brown!10} + Downsample & \textbf{0.0} {\scriptsize (100.0)} & \textbf{0.0} {\scriptsize (100.0)} & \textbf{0.0} {\scriptsize (100.0)} & \textbf{0.0} {\scriptsize (100.0)} & \textbf{0.0} {\scriptsize (100.0)} & \textbf{0.0} {\scriptsize (100.0)} & \textbf{0.0} {\scriptsize (100.0)} & \textbf{0.0} {\scriptsize (100.0)} & \textbf{0.0} {\scriptsize (100.0)} \\ \midrule

+ FastV~\cite{chen2024image} & 
\makecell{\textbf{57.8} {\scriptsize (87.6)} \\ \deltayellow{29.8}} & \makecell{\textbf{45.2} {\scriptsize (95.9)} \\ \deltayellow{50.7}} & \makecell{\textbf{56.5} {\scriptsize (95.8)} \\ \deltayellow{39.3}} & \makecell{\textbf{78.9} {\scriptsize (96.7)} \\ \deltayellow{17.8}} & \makecell{\textbf{65.4} {\scriptsize (94.8)} \\ \deltayellow{29.4}} & \makecell{\textbf{41.0} {\scriptsize (76.0)} \\ \deltayellow{35.0}} & \makecell{\textbf{29.1} {\scriptsize (57.2)} \\ \deltayellow{28.1}} & \makecell{\textbf{35.0} {\scriptsize (78.1)} \\ \deltayellow{43.1}} & \makecell{\textbf{51.1} {\scriptsize (85.3)} \\ \deltayellow{34.2}} \\ \cmidrule(lr){2-10}

+ VisionZip~\cite{yang2025visionzip} & 
\makecell{\textbf{59.3} {\scriptsize (91.2)} \\ \deltayellow{31.9}} & \makecell{\textbf{42.4} {\scriptsize (93.8)} \\ \deltayellow{51.4}} & \makecell{\textbf{42.2} {\scriptsize (93.6)} \\ \deltayellow{51.4}} & \makecell{\textbf{54.9} {\scriptsize (95.3)} \\ \deltayellow{40.4}} & \makecell{\textbf{72.5} {\scriptsize (96.8)} \\ \deltayellow{24.3}} & \makecell{\textbf{45.9} {\scriptsize (81.4)} \\ \deltayellow{35.5}} & \makecell{\textbf{29.6} {\scriptsize (58.1)} \\ \deltayellow{28.5}} & \makecell{\textbf{51.2} {\scriptsize (87.3)} \\ \deltayellow{36.1}} & \makecell{\textbf{49.8} {\scriptsize (87.2)} \\ \deltayellow{37.4}} \\ \cmidrule(lr){2-10}

+ PruMerge+~\cite{shang2024llava} & 
\makecell{\textbf{57.7} {\scriptsize (91.9)} \\ \deltayellow{34.2}} & \makecell{\textbf{51.2} {\scriptsize (95.1)} \\ \deltayellow{43.9}} & \makecell{\textbf{52.6} {\scriptsize (94.6)} \\ \deltayellow{42.0}} & \makecell{\textbf{62.0} {\scriptsize (95.9)} \\ \deltayellow{33.9}} & \makecell{\textbf{72.1} {\scriptsize (97.5)} \\ \deltayellow{25.4}} & \makecell{\textbf{48.1} {\scriptsize (82.3)} \\ \deltayellow{34.2}} & \makecell{\textbf{21.2} {\scriptsize (46.2)} \\ \deltayellow{25.0}} & \makecell{\textbf{40.5} {\scriptsize (73.6)} \\ \deltayellow{33.1}} & \makecell{\textbf{50.7} {\scriptsize (84.6)} \\ \deltayellow{33.9}} \\ \cmidrule(lr){2-10}

+ DART~\cite{wen2025stop} & 
\makecell{\textbf{58.9} {\scriptsize (88.1)} \\ \deltayellow{29.2}} & \makecell{\textbf{54.8} {\scriptsize (94.9)} \\ \deltayellow{40.1}} & \makecell{\textbf{52.2} {\scriptsize (94.6)} \\ \deltayellow{42.4}} & \makecell{\textbf{67.6} {\scriptsize (94.9)} \\ \deltayellow{27.3}} & \makecell{\textbf{69.4} {\scriptsize (94.5)} \\ \deltayellow{25.1}} & \makecell{\textbf{47.0} {\scriptsize (77.7)} \\ \deltayellow{30.7}} & \makecell{\textbf{40.2} {\scriptsize (70.2)} \\ \deltayellow{30.0}} & \makecell{\textbf{39.0} {\scriptsize (69.0)} \\ \deltayellow{30.0}} & \makecell{\textbf{53.6} {\scriptsize (85.5)} \\ \deltayellow{31.9}} \\ \midrule

% --- Section 2: 88.89% ---
\rowcolor{lightgrayblue}  \textit{\textbf{Group A} {\scriptsize (Group B)}} & \multicolumn{9}{c}{\textit{Token Reduction ($\downarrow$ 88.89\%)}} \\ \midrule
\rowcolor{brown!10} + Downsample & \textbf{0.0} {\scriptsize (100.0)} & \textbf{0.0} {\scriptsize (100.0)} & \textbf{0.0} {\scriptsize (100.0)} & \textbf{0.0} {\scriptsize (100.0)} & \textbf{0.0} {\scriptsize (100.0)} & \textbf{0.0} {\scriptsize (100.0)} & \textbf{0.0} {\scriptsize (100.0)} & \textbf{0.0} {\scriptsize (100.0)} & \textbf{0.0} {\scriptsize (100.0)} \\ \midrule

+ FastV~\cite{chen2024image} & 
\makecell{\textbf{44.5} {\scriptsize (82.5)} \\ \deltayellow{38.0}} & \makecell{\textbf{39.2} {\scriptsize (90.3)} \\ \deltayellow{51.1}} & \makecell{\textbf{44.1} {\scriptsize (90.8)} \\ \deltayellow{46.7}} & \makecell{\textbf{59.4} {\scriptsize (94.0)} \\ \deltayellow{34.6}} & \makecell{\textbf{46.8} {\scriptsize (88.7)} \\ \deltayellow{41.9}} & \makecell{\textbf{31.0} {\scriptsize (73.0)} \\ \deltayellow{42.0}} & \makecell{\textbf{17.8} {\scriptsize (41.3)} \\ \deltayellow{23.5}} & \makecell{\textbf{28.4} {\scriptsize (61.7)} \\ \deltayellow{33.3}} & \makecell{\textbf{38.9} {\scriptsize (77.8)} \\ \deltayellow{38.9}} \\ \cmidrule(lr){2-10}

+ VisionZip~\cite{yang2025visionzip} & 
\makecell{\textbf{49.4} {\scriptsize (83.4)} \\ \deltayellow{34.0}} & \makecell{\textbf{33.2} {\scriptsize (89.0)} \\ \deltayellow{55.8}} & \makecell{\textbf{44.4} {\scriptsize (88.1)} \\ \deltayellow{43.7}} & \makecell{\textbf{48.1} {\scriptsize (92.2)} \\ \deltayellow{44.1}} & \makecell{\textbf{70.0} {\scriptsize (92.3)} \\ \deltayellow{22.3}} & \makecell{\textbf{30.3} {\scriptsize (73.0)} \\ \deltayellow{42.7}} & \makecell{\textbf{22.0} {\scriptsize (36.4)} \\ \deltayellow{14.4}} & \makecell{\textbf{49.7} {\scriptsize (74.4)} \\ \deltayellow{24.7}} & \makecell{\textbf{43.4} {\scriptsize (78.6)} \\ \deltayellow{35.2}} \\ \cmidrule(lr){2-10}

+ PruMerge+~\cite{shang2024llava} & 
\makecell{\textbf{50.4} {\scriptsize (85.8)} \\ \deltayellow{35.4}} & \makecell{\textbf{36.9} {\scriptsize (87.2)} \\ \deltayellow{50.3}} & \makecell{\textbf{38.4} {\scriptsize (86.4)} \\ \deltayellow{48.0}} & \makecell{\textbf{42.9} {\scriptsize (91.9)} \\ \deltayellow{49.0}} & \makecell{\textbf{71.5} {\scriptsize (94.2)} \\ \deltayellow{22.7}} & \makecell{\textbf{28.8} {\scriptsize (71.6)} \\ \deltayellow{42.8}} & \makecell{\textbf{18.1} {\scriptsize (33.0)} \\ \deltayellow{14.9}} & \makecell{\textbf{43.5} {\scriptsize (73.8)} \\ \deltayellow{30.3}} & \makecell{\textbf{41.3} {\scriptsize (78.0)} \\ \deltayellow{36.7}} \\ \cmidrule(lr){2-10}

+ DART~\cite{wen2025stop} & 
\makecell{\textbf{47.5} {\scriptsize (81.2)} \\ \deltayellow{33.7}} & \makecell{\textbf{40.5} {\scriptsize (87.7)} \\ \deltayellow{47.2}} & \makecell{\textbf{40.9} {\scriptsize (86.9)} \\ \deltayellow{46.0}} & \makecell{\textbf{49.6} {\scriptsize (91.7)} \\ \deltayellow{42.1}} & \makecell{\textbf{57.7} {\scriptsize (90.9)} \\ \deltayellow{33.2}} & \makecell{\textbf{35.4} {\scriptsize (70.0)} \\ \deltayellow{34.6}} & \makecell{\textbf{31.5} {\scriptsize (63.2)} \\ \deltayellow{31.7}} & \makecell{\textbf{27.3} {\scriptsize (57.6)} \\ \deltayellow{30.3}} & \makecell{\textbf{41.3} {\scriptsize (78.6)} \\ \deltayellow{37.3}} \\ \midrule

% --- Section 3: 93.75% ---
\rowcolor{lightgrayblue}  \textit{\textbf{Group A} {\scriptsize (Group B)}} & \multicolumn{9}{c}{\textit{Token Reduction ($\downarrow$ 93.75\%)}} \\ \midrule
\rowcolor{brown!10} + Downsample & \textbf{0.0} {\scriptsize (100.0)} & \textbf{0.0} {\scriptsize (100.0)} & \textbf{0.0} {\scriptsize (100.0)} & \textbf{0.0} {\scriptsize (100.0)} & \textbf{0.0} {\scriptsize (100.0)} & \textbf{0.0} {\scriptsize (100.0)} & \textbf{0.0} {\scriptsize (100.0)} & \textbf{0.0} {\scriptsize (100.0)} & \textbf{0.0} {\scriptsize (100.0)} \\ \midrule

+ FastV~\cite{chen2024image} & 
\makecell{\textbf{35.7} {\scriptsize (81.4)} \\ \deltayellow{45.7}} & \makecell{\textbf{31.9} {\scriptsize (85.7)} \\ \deltayellow{53.8}} & \makecell{\textbf{35.3} {\scriptsize (86.6)} \\ \deltayellow{51.3}} & \makecell{\textbf{48.8} {\scriptsize (91.5)} \\ \deltayellow{42.7}} & \makecell{\textbf{37.4} {\scriptsize (88.1)} \\ \deltayellow{50.7}} & \makecell{\textbf{22.8} {\scriptsize (74.5)} \\ \deltayellow{51.7}} & \makecell{\textbf{13.3} {\scriptsize (33.0)} \\ \deltayellow{19.7}} & \makecell{\textbf{14.8} {\scriptsize (74.8)} \\ \deltayellow{60.0}} & \makecell{\textbf{30.0} {\scriptsize (77.0)} \\ \deltayellow{47.0}} \\ \cmidrule(lr){2-10}

+ VisionZip~\cite{yang2025visionzip} & 
\makecell{\textbf{41.0} {\scriptsize (79.0)} \\ \deltayellow{38.0}} & \makecell{\textbf{34.5} {\scriptsize (81.9)} \\ \deltayellow{47.4}} & \makecell{\textbf{33.3} {\scriptsize (82.2)} \\ \deltayellow{48.9}} & \makecell{\textbf{43.5} {\scriptsize (88.4)} \\ \deltayellow{44.9}} & \makecell{\textbf{66.3} {\scriptsize (89.4)} \\ \deltayellow{23.1}} & \makecell{\textbf{24.3} {\scriptsize (69.8)} \\ \deltayellow{45.5}} & \makecell{\textbf{14.0} {\scriptsize (25.2)} \\ \deltayellow{11.2}} & \makecell{\textbf{26.1} {\scriptsize (71.3)} \\ \deltayellow{45.2}} & \makecell{\textbf{35.4} {\scriptsize (73.4)} \\ \deltayellow{38.0}} \\ \cmidrule(lr){2-10}

+ PruMerge+~\cite{shang2024llava} & 
\makecell{\textbf{43.0} {\scriptsize (76.7)} \\ \deltayellow{33.7}} & \makecell{\textbf{29.6} {\scriptsize (76.9)} \\ \deltayellow{47.3}} & \makecell{\textbf{34.1} {\scriptsize (76.1)} \\ \deltayellow{42.0}} & \makecell{\textbf{43.0} {\scriptsize (87.8)} \\ \deltayellow{44.8}} & \makecell{\textbf{67.7} {\scriptsize (87.6)} \\ \deltayellow{19.9}} & \makecell{\textbf{25.5} {\scriptsize (65.5)} \\ \deltayellow{40.0}} & \makecell{\textbf{12.6} {\scriptsize (21.8)} \\ \deltayellow{9.2}} & \makecell{\textbf{29.4} {\scriptsize (68.9)} \\ \deltayellow{39.5}} & \makecell{\textbf{35.6} {\scriptsize (70.2)} \\ \deltayellow{34.6}} \\ \cmidrule(lr){2-10}

+ DART~\cite{wen2025stop} & 
\makecell{\textbf{41.9} {\scriptsize (78.8)} \\ \deltayellow{36.9}} & \makecell{\textbf{33.8} {\scriptsize (81.8)} \\ \deltayellow{48.0}} & \makecell{\textbf{38.4} {\scriptsize (80.4)} \\ \deltayellow{42.0}} & \makecell{\textbf{46.9} {\scriptsize (88.9)} \\ \deltayellow{42.0}} & \makecell{\textbf{57.0} {\scriptsize (88.5)} \\ \deltayellow{31.5}} & \makecell{\textbf{26.2} {\scriptsize (61.8)} \\ \deltayellow{35.6}} & \makecell{\textbf{25.6} {\scriptsize (57.4)} \\ \deltayellow{31.8}} & \makecell{\textbf{14.5} {\scriptsize (67.1)} \\ \deltayellow{52.6}} & \makecell{\textbf{35.5} {\scriptsize (75.6)} \\ \deltayellow{40.1}} \\ 

\bottomrule
\end{tabular}}
\end{table*}

\subsection{Experiments Setup}
Before conducting experiments, it is crucial to choose a suitable MLLM for achieving the downsampling method. 
Most MLLMs only support fixed-resolution inputs, which makes it impossible to achieve the downsampling method. 
In other words, for such MLLMs, no matter which resolution the original image is downsampled to, the image will finally be resized to a fixed resolution, making the downsampling meaningless. 
Considering that Qwen2-VL~\cite{wang2024qwen2} and Qwen2.5-VL~\cite{bai2025qwen2}, based on the naive dynamic resolution mechanism and M-RoPE techniques, are the open-source MLLMs closest to realizing the concept of allowing arbitrary resolution inputs, we choose Qwen2-VL\footnote{When conducting this experiment, on one hand, there is no repository that systematically implements visual token compression method based on Qwen2.5-VL; on the other hand, there are few open-source visual token compression methods adapted for Qwen2.5-VL. Therefore, we use Qwen2-VL as the main experimental model.} in our comparison experiments, which supports the downsampling method the best.
In order to ensure that downsampling occurs at the original resolution as much as possible without adding extra resizing operations, we set Qwen2-VL's max pixels and min pixels to 2408448 and 3136. 
In this case, only a few extremely high-resolution images are resized before downsampling to ensure sufficient GPU memory.

To guarantee comprehensive experiments, we select four typical token compression methods%\footnote{Results of DART are reproduced based on \hyperlink{https://github.com/ZichenWen1/DART}{https://github.com/ZichenWen1/DART}. Results of FastV, VisionZip, and PruMerge+ are reproduced based on \hyperlink{https://github.com/EffiVLM-Bench/EffiVLM-Bench}{https://github.com/EffiVLM-Bench/EffiVLM-Bench}.} 
: FastV~\cite{chen2024image}, VisionZip~\cite{yang2025visionzip}, PruMerge+~\cite{shang2024llava}, and DART~\cite{wen2025stop}, with the token compression ratio set to: 75.00\%, 88.89\%, 93.75\%, 96.00\%, and 99.00\%. 
For the token compression ratio $C$, the downsampling method applies an equivalent downsampling ratio $D$ for fairness.
The rule is shown in Eq.~\ref{Equ1}. 
Moreover, we choose eight popular benchmarks, including six general benchmarks (GQA~\cite{hudson2019gqa}, MMBench\_EN~\cite{liu2024mmbench}, MMBench\_CN~\cite{liu2024mmbench}, MME~\cite{yin2024survey}, POPE~\cite{li2023evaluating}, and MMStar~\cite{chen2024we}) and two resolution-sensitive OCR benchmarks (OCRBench~\cite{liu2024ocrbench}, and ChartQA~\cite{masry2022chartqa}).

\begin{equation}
\setlength{\abovedisplayskip}{3pt}
\setlength{\belowdisplayskip}{3pt}
\label{Equ1}
    \frac{1}{D^2} \times 100\%= 1-C
\end{equation}

\subsection{Results Analysis}
\textbf{Comparison between different methods:} 
Across diverse compression ratios and general-purpose benchmarks, simple image downsampling consistently outperforms sophisticated token compression methods. 
For instance, at 93.75\% compression, downsampling achieves 66.4\% on MMBench, outperforming the best advanced method, DART, by a 24.3\% relative improvement.
Meanwhile, with a constraint of 70\% of vanilla performance, downsampling can achieve a compression ratio of 93.75\%, while other methods can only meet this condition at a compression ratio of 88.89\%. 
The results verify a basic phenomenon in the field of visual token compression: \textbf{\textit{a substantial portion of samples in general-purpose benchmarks can be correctly answered using only low-resolution global information, without requiring the fine-grained visual details that advanced methods strive to preserve.}}

\textbf{Comparison between different compression ratios:} 
As compression becomes increasingly aggressive (96.00\% and 99.00\%), all sophisticated token compression methods experience performance degradation, while image downsampling demonstrates remarkably graceful degradation. 
At 99.00\% compression, downsampling maintains a score of 51.6\% on MMBench, while FastV and PruMerge+ decrease to approximately 24\%. 
The results further verify the phenomenon above: \textit{\textbf{in the existing general-purpose benchmarks, image downsampling can fully meet the acceleration requirements for most samples.}}

\textbf{Comparison between different tasks:} 
On tasks requiring fine-grained visual understanding, particularly chart comprehension, we observe a reversal of the phenomenon mentioned above. 
At moderate compression ratios (93.75\% and 88.89\%), VisionZip and FastV outperform image downsampling on ChartQA by significant margins. 
This divergence is highly informative: while image downsampling uniformly preserves global information at the expense of local details, the sophisticated compression methods can selectively retain text regions and numeric values that are critical for chart understanding, which can be considered difficult to compress.
Thus, a deeper observation of the above phenomenon can be concluded:\textit{\textbf{the sophisticated token compression methods demonstrate the expected effectiveness in tasks that require fine-grained visual understanding.}}

The comparisons across methods, compression ratios, and tasks provide compelling evidence that current benchmarks contain a substantial simplicity bias.
The performance advantage of image downsampling emerges not from its sophistication but from its ability to adequately address samples that don't require fine-grained visual understanding, precisely the samples that dominate current benchmarks.
Thus, based on the experimental results and the comparisons, we propose a well-founded hypothesis in Section~\ref{hypothesis}.

\begin{figure*}
    \centering
    \includegraphics[width=\linewidth]{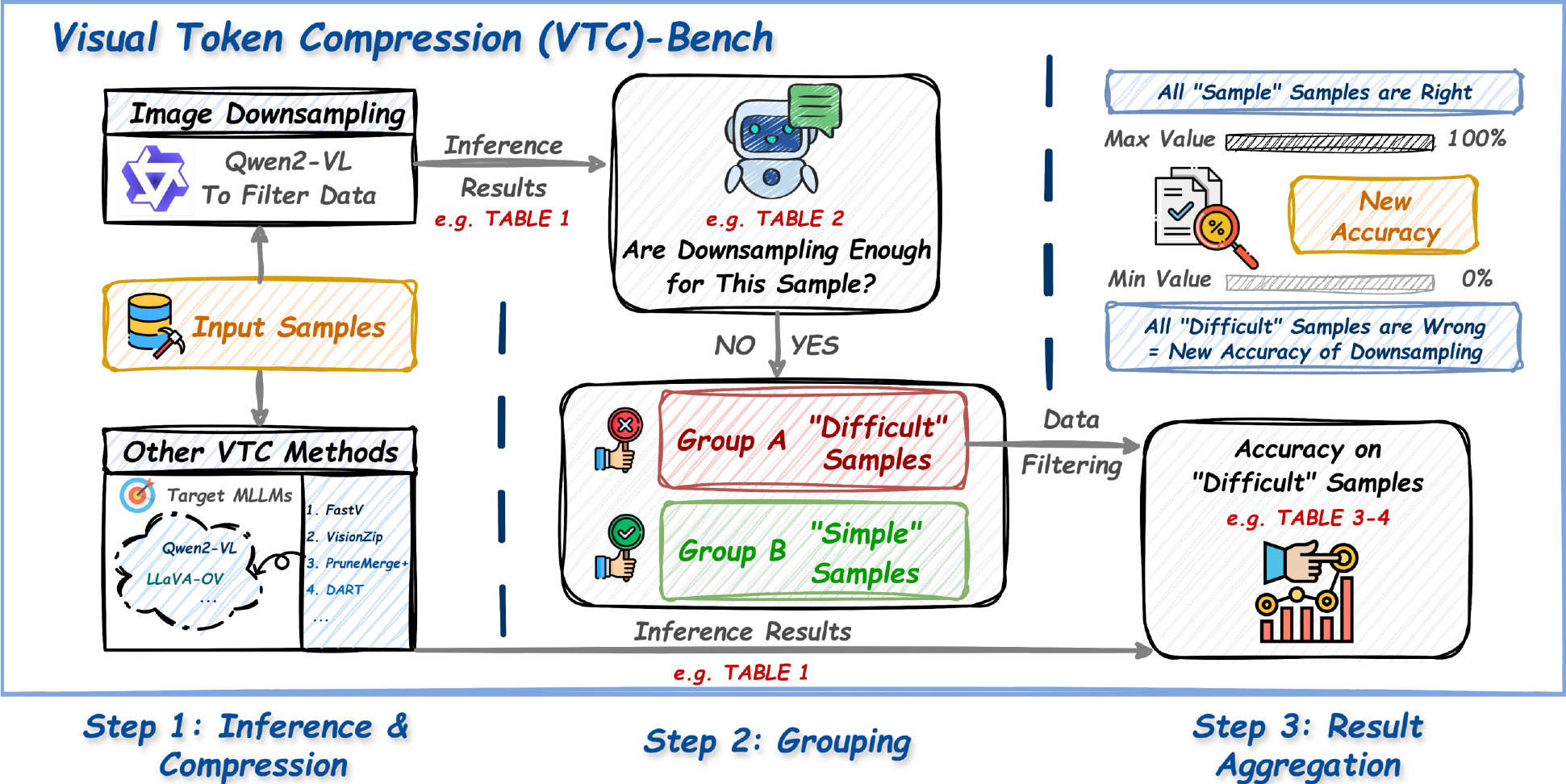}
    \caption{The VTC-Bench is a simple but effective framework that can transform any existing benchmarks to a subset that can fairly evaluate VTC (Visual Token Compression) methods. The samples that are answered correctly by the original Qwen2-VL model without downsampling form the input samples. More details in Section~\ref{Framwork}. }
    \label{figuremain}
\end{figure*}

\subsection{Hypothesis}
\label{hypothesis}

%In real life, if the difficulty of an exam is much lower than that of students, then students' grades will be chaotic, mainly manifested in the confusion of good students' and bad students' grades.
In the field of visual token compression, there is a general reliance on existing benchmarks, without ever considering whether these data are suitable for the visual token compression task. 
Based on the above experimental results in Table \ref{table1}, we propose a bold hypothesis: 
\textbf{\textit{Some data in the existing benchmarks is overly simplistic and irrelevant to evaluating visual token compression methods, leading to the unreasonable phenomenon that even the downsampling method is sufficient to deal with the visual token compression task}}. 

To validate this hypothesis, we design a data-centric analysis using downsampling as a discriminator.
We first drop out the samples answered incorrectly at the original resolution, which we consider are too hard for the original models to understand, not to mention the compressed models. Then, for a given compression ratio, we classify each sample in a benchmark into one of two groups based on the performance of the downsampling method:
\ding{172} Difficult Samples (Group A): Samples that are answered incorrectly by the downsampling method. \ding{173} Simple Samples (Group B): Samples that are answered correctly by the downsampling method.
We then evaluate all compression methods on these two groups separately to assess whether the sophisticated methods demonstrate their expected superiority on the “difficult” samples where image downsampling fails.
Results from Table~\ref{table2} strongly confirm our hypothesis, followed by two key conclusions as follows.

\textbf{\ding{172} Significant performance gap between groups:}
Across all benchmarks and compression methods, the accuracy on “simple” samples (Group B) is dramatically higher than on “difficult” samples (Group A). 
For instance, on GQA at 75\% compression, the accuracy of all methods on simple samples is above 87.6\%, while on difficult samples, it drops to a maximum of 59.3\% (VisionZip). 
This stark contrast is common in Table \ref{table2}, demonstrating that the two groups represent essentially different levels of visual comprehension difficulty.
The existence of this gap validates our core hypothesis that \textbf{\textit{the benchmark comprises a mixture of ``simple'' and ``difficult'' samples. 
In other words, the current benchmarks are noisy for evaluating the visual token compression methods, where the noise refers to the task-irrelevant samples.}}
Moreover, the significant gap also proves that \textbf{\textit{downsampling can serve as a clever filter to distinguish between ``simple'' and ``difficult'' samples, which can be the key to denoise the current benchmarks.}}

\textbf{\ding{173} Ideal reference points brought by downsampling:}
The 0\%/100\% dichotomy created by image downsampling provides ideal reference points for evaluation.
In Group B, where downsampling achieves 100\% accuracy, advanced methods show comparable but not superior performance (e.g., 87.6-91.9\% on GQA at 75\% compression), confirming that their sophisticated approaches offer no advantage for simple samples.
In Group A, where downsampling fails completely (0\% accuracy), advanced methods demonstrate their true value by significantly exceeding this baseline. For instance, DART achieves 40.2\% on OCRBench and VisionZip reaches 51.2\% on ChartQA at 75\% compression, proving their ability to preserve crucial details that downsampling loses.
\begin{table*}[!h]
\small
\renewcommand{\arraystretch}{0.9}
\centering
\caption{VTC-Bench results on Qwen2-VL-7B. Take Qwen2-VL-7B as the data filter model. }
\label{table3}\renewcommand{\tabcolsep}{8pt}
\resizebox{\textwidth}{!}{
\begin{tabular}{l|c c c c c c c c |c}
\toprule
\textit{\textbf{Method}} & \textit{\textbf{GQA} }& \textit{\textbf{MMB}} & \textit{\textbf{MMB$^{CN}$} }& \textit{\textbf{MME}} & \textit{\textbf{POPE}} & \textit{\textbf{MMStar}}  & \textit{\textbf{OCRBench}}& \textit{\textbf{ChartQA}}  & \textit{\textbf{Average}} \\
\midrule
\rowcolor{lightgrayblue} \textit{Qwen2-VL-7B~\cite{wang2024qwen2}} & \multicolumn{9}{c}{\textit{Token Reduction ($\downarrow$ 75.00\%)}} \\ \midrule
 
+ Downsample (Baseline) & 0.0 & 0.0 & 0.0 & 0.0 & 0.0 & 0.0 & 0.0 & 0.0 & 0.0 \\
\rowcolor{gray!10}+ FastV~\cite{chen2024image} & 57.8 & 45.2 & \textbf{56.5} & \underline{78.9} & 65.4 & 41.0 & 29.1 & 35.0 & \underline{51.1} \\
 + VisionZip~\cite{yang2025visionzip} & \textbf{59.3} & 42.4 & 42.2 & 54.9 & \textbf{72.5} & 45.9 & \underline{29.6} & \textbf{51.2} & 49.8 \\
\rowcolor{gray!10}+ PruMerge+~\cite{shang2024llava} & 57.7 & \underline{51.2} & \underline{52.6} & 62.0 & \underline{72.1} & \textbf{48.1} & 21.2 & \underline{40.5} & 50.7 \\
 + DART~\cite{wen2025stop} & \underline{58.9} & \textbf{54.8} & 52.2 & \textbf{67.6} & 69.4 & \underline{47.0} & \textbf{40.2} & 39.0 & \textbf{53.6} \\
\midrule
\rowcolor{lightgrayblue} \textit{Qwen2-VL-7B~\cite{wang2024qwen2}} & \multicolumn{9}{c}{\textit{Token Reduction ($\downarrow$ 88.89\%)}} \\ \midrule
 + Downsample (Baseline)& 0.0 & 0.0 & 0.0 & 0.0 & 0.0 & 0.0 & 0.0 & 0.0 & 0.0 \\
\rowcolor{gray!10} + FastV~\cite{chen2024image} & 44.5 & \underline{39.2} & \underline{44.1} & \textbf{59.4} & 46.8 & \underline{31.0} & 17.8 & 28.4 & 38.9 \\
 + VisionZip~\cite{yang2025visionzip}  & \underline{49.4} & 33.2 & \textbf{44.4} & 48.1 & \underline{70.0} & 30.3 & \underline{22.0} & \textbf{49.7} & \textbf{43.4} \\
\rowcolor{gray!10}+ PruMerge+~\cite{shang2024llava} & \textbf{50.4} & 36.9 & 38.4 & 42.9 & \textbf{71.5} & 28.8 & 18.1 & \underline{43.5} & \underline{41.3} \\
 + DART~\cite{wen2025stop}& 47.5 & \textbf{40.5} & 40.9 & \underline{49.6} & 57.7 & \textbf{35.4} & \textbf{31.5} & 27.3 & \underline{41.3} \\
\midrule
\rowcolor{lightgrayblue} \textit{Qwen2-VL-7B~\cite{wang2024qwen2}} & \multicolumn{9}{c}{\textit{Token Reduction ($\downarrow$ 93.75\%)}} \\ \midrule
 
+ Downsample (Baseline) & 0.0 & 0.0 & 0.0 & 0.0 & 0.0 & 0.0 & 0.0 & 0.0 & 0.0 \\
\rowcolor{gray!10}+ FastV~\cite{chen2024image} & 35.7 & 31.9 & \underline{35.3} & \textbf{48.8} & 37.4 & 22.8 & 13.3 & 14.8 & 30.0 \\
 + VisionZip~\cite{yang2025visionzip}  & 41.0 & \textbf{34.5} & 33.3 & 43.5 & \underline{66.3} & 24.3 & \underline{14.0} & \underline{26.1} & 35.4 \\
\rowcolor{gray!10}+ PruMerge+~\cite{shang2024llava} & \textbf{43.0} & 29.6 & 34.1 & 43.0 & \textbf{67.7} & \underline{25.5} & 12.6 & \textbf{29.4} & \textbf{35.6} \\
 + DART~\cite{wen2025stop}& \underline{41.9} & \underline{33.8} & \textbf{38.4} & \underline{46.9} & 57.0 & \textbf{26.2} & \textbf{25.6} & 14.5 & \underline{35.5} \\
\midrule
\rowcolor{lightgrayblue} \textit{Qwen2-VL-7B~\cite{wang2024qwen2}} & \multicolumn{9}{c}{\textit{Token Reduction ($\downarrow$ 96.00\%)}} \\ \midrule
 
+ Downsample (Baseline) & 0.0 & 0.0 & 0.0 & 0.0 & 0.0 & 0.0 & 0.0 & 0.0 & 0.0 \\
\rowcolor{gray!10}+ FastV~\cite{chen2024image} & 29.6 & 24.7 & \underline{33.1} & 35.6 & 35.6 & 21.5 & \underline{11.0} & 9.3 & 25.0 \\
 + VisionZip~\cite{yang2025visionzip} & \underline{38.6} & \underline{31.2} & 32.6 & \textbf{37.9} & \textbf{60.0} & \textbf{24.5} & \underline{11.0} & \underline{15.1} & \underline{31.4} \\
\rowcolor{gray!10}+ PruMerge+~\cite{shang2024llava} & \textbf{38.8} & 26.0 & 29.3 & 37.0 & \underline{56.4} & \underline{22.6} & 9.2 & \textbf{17.0} & 29.5 \\
 + DART~\cite{wen2025stop}& 36.4 & \textbf{33.7} & \textbf{36.4} & \textbf{37.9} & 53.2 & 22.3 & \textbf{24.7} & 10.5 & \textbf{31.9} \\
\midrule
\rowcolor{lightgrayblue} \textit{Qwen2-VL-7B~\cite{wang2024qwen2}} & \multicolumn{9}{c}{\textit{Token Reduction ($\downarrow$ 99.00\%)}} \\ \midrule
+ Downsample (Baseline) & 0.0 & 0.0 & 0.0 & 0.0 & 0.0 & 0.0 & 0.0 & 0.0 & 0.0 \\
\rowcolor{gray!10} + FastV~\cite{chen2024image} & 18.3 & 18.3 & 21.5 & 21.5 & 44.3 & 15.0 & \underline{4.2} & 3.8 & 18.4 \\
+ VisionZip~\cite{yang2025visionzip} & \underline{23.4} & \textbf{28.8} & \textbf{32.2} & \underline{28.5} & \textbf{53.6} & \textbf{19.4} & 3.7 & \underline{5.5} & \underline{24.4} \\
\rowcolor{gray!10} + PruMerge+~\cite{shang2024llava} & 20.7 & 17.8 & 21.1 & 22.9 & \underline{52.6} & 17.1 & 2.5 & \textbf{7.1} & 20.2 \\
+ DART~\cite{wen2025stop} & \textbf{24.5} & \underline{26.5} & \underline{28.1} & \textbf{30.6} & 41.5 & \underline{19.2} & \textbf{25.6} & 4.2 & \textbf{25.0} \\
\bottomrule
\end{tabular}}
\end{table*}
\begin{table}[t]
\scriptsize
\centering
\renewcommand{\arraystretch}{0.8}
\caption{VTC-Bench results on LLaVA-OV-7B. Take Qwen2-VL-7B as the data filter model.}
\label{table4}
\renewcommand{\tabcolsep}{4pt}
\resizebox{0.49\textwidth}{!}{
\begin{tabular}{l|c c c c c}
\toprule
\textit{\textbf{Method}} & \textit{\textbf{GQA}} & \textit{\textbf{MMB}} & \textit{\textbf{MMB$^{CN}$}} & \textit{\textbf{POPE}} & \textbf{\textit{MMStar}} \\
\midrule
\rowcolor{lightgrayblue} \textit{LLaVA-OV-7B~\cite{li2024llava}} & \multicolumn{5}{c}{\textit{Token Reduction ($\downarrow$ 75.00\%)}} \\
 \midrule
+ FastV~\cite{chen2024image} & 54.3 & \underline{70.5} & 69.1 & 63.8 & \textbf{48.6}  \\
\rowcolor{gray!10}+ VisionZip~\cite{yang2025visionzip} & \underline{59.0} & 67.7 & \underline{71.3} & \textbf{80.8} & 44.8\\
+ PruMerge+~\cite{shang2024llava} & \textbf{60.4} & \textbf{74.2} & \textbf{73.5} & \underline{75.6} & \textbf{48.6}\\
\midrule
\rowcolor{lightgrayblue}\textit{LLaVA-OV-7B~\cite{li2024llava}} & \multicolumn{5}{c}{\textit{Token Reduction ($\downarrow$ 88.89\%)}} \\
 \midrule
+ FastV~\cite{chen2024image} & 45.3 & 64.6 & 66.4 & 39.1 & 42.4\\
\rowcolor{gray!10}+ VisionZip~\cite{yang2025visionzip} & \underline{56.6} & \textbf{71.9} & \underline{71.2} & \underline{69.6} & \underline{43.5}\\
+ PruMerge+~\cite{shang2024llava} & \textbf{57.4} & \underline{68.8} & \textbf{71.5} & \textbf{76.0} & \textbf{45.8}\\
\midrule
\rowcolor{lightgrayblue} \textit{LLaVA-OV-7B~\cite{li2024llava}} & \multicolumn{5}{c}{\textit{Token Reduction ($\downarrow$ 93.75\%)}} \\
 \midrule
+ FastV~\cite{chen2024image} & 36.7 & 51.2 & 53.3 & 29.7 & 32.6\\
\rowcolor{gray!10}+ VisionZip~\cite{yang2025visionzip} & \underline{49.1} & \underline{64.3} & \underline{62.4} & \underline{53.6} & \textbf{36.6}\\
+ PruMerge+~\cite{shang2024llava} & \textbf{50.2} & \textbf{66.6} & \textbf{65.3} & \textbf{59.9} & \underline{34.8}\\
\midrule
\rowcolor{lightgrayblue} \textit{LLaVA-OV-7B~\cite{li2024llava}} & \multicolumn{5}{c}{\textit{Token Reduction ($\downarrow$ 96.00\%)}} \\
 \midrule
+ FastV~\cite{chen2024image} & 31.4 & 37.4 & 43.1 & 24.5 & 28.6\\
\rowcolor{gray!10}+ VisionZip~\cite{yang2025visionzip} & \underline{42.6} & \underline{55.4} & \underline{56.9} & \underline{45.4} & \underline{30.8}\\
+ PruMerge+~\cite{shang2024llava} & \textbf{42.7} & \textbf{57.8} & \textbf{59.6} & \textbf{49.9} & \textbf{31.3}\\
\midrule
\rowcolor{lightgrayblue} \textit{LLaVA-OV-7B~\cite{li2024llava}} & \multicolumn{5}{c}{\textit{Token Reduction ($\downarrow$ 99.00\%)}} \\
 \midrule
+ FastV~\cite{chen2024image} & \underline{25.7} & \underline{25.8} & \underline{29.6} & 39.3 & 21.9\\
\rowcolor{gray!10}+ VisionZip~\cite{yang2025visionzip} & \textbf{28.3} & \textbf{28.1} & \textbf{32.8} & \textbf{42.1} & \underline{24.7}\\
+ PruMerge+~\cite{shang2024llava} & 25.3 & 25.5 & 28.5 & \underline{40.4} & \textbf{25.2}\\
\bottomrule
\end{tabular}}
\end{table}

\subsection{Summary}
In this section, we conduct two comprehensive experiments to further understand the anomalous phenomenon: image downsampling exceeds other sophisticated methods under some settings. 
The first experiment validates the universality of this anomalous phenomenon and introduces our basic hypothesis:
Some data in the existing benchmarks is overly simplistic, leading to the unreasonable phenomenon that even the simplest downsampling method is sufficient to deal with the visual token compression task.
Furthermore, the second experiment further validates this hypothesis and proves that the current benchmarks are noisy for evaluating the visual token compression methods. Moreover, the second experiment simultaneously demonstrates that downsampling can serve as a clever filter to distinguish between ``simple'' and ``difficult'' samples, which can be the key to denoise the current benchmarks.

%% file: sections/4_bench.tex
\section{Evaluation Framework}

\subsection{Framework Construction}
\label{Framwork}
To address the simplicity bias and denoise existing benchmarks for the visual token compression task, we propose the VTC-Bench (Visual Token Compression Benchmark) framework, a novel framework specifically designed for the fair and effective evaluation of visual token compression methods. 
The construction is based on the key insight, validated in Section~\ref{hypothesis}, that ``downsampling can serve as a clever filter to distinguish between `simple' and `difficult' samples''. 
We leverage this idea to construct a challenging benchmark comprising predominantly “difficult” samples that require fine-grained visual understanding.
This process, summarized in Figure \ref{figuremain}, does not create new data but rather applies a rigorous filtering mechanism to existing benchmarks to identify a challenging evaluation and noise-free set. 
The pipeline consists of three critical steps executed for each candidate sample and dynamically adapts to different compression ratios:

% \noindent \textbf{Step 1: Inference \& Compression:}
% For a given sample and a target token compression ratio, the models are inferred by two types of methods: 
% \ding{172} The downsampling method (the filter), which applies an equivalent downsampling ratio calculated by Eq.~\ref{Equ1} for a fair comparison. \textbf{(Based on Qwen2-VL that has similar parameters to the target MLLM) }
% \ding{173} Other sophisticated visual token compression methods, which await evaluation (e.g., FastV, VisionZip, DART). 
% This step not only prepares for the subsequent evaluation of visual token compression methods but also provides a basis for the subsequent sample filtering. \textbf{(Based on the target MLLM)} 
\paragraph{Step 1: Inference \& Compression:}
Given a sample and a target token compression ratio, we run two inference pipelines: \ding{172} a downsampling baseline (the filter) including one model that applies the equivalent ratio from Eq.~\ref{Equ1} for a fair comparison and another original model without downsampling, implemented with Qwen2-VL which has a similar number of parameter with the target MLLM; and \ding{173} advanced visual token compression methods (e.g., FastV, VisionZip, DART) evaluated directly on the target MLLM. This step both establishes a fair basis for assessing compression approaches and provides signals for subsequent sample filtering.

\paragraph{Step 2: Grouping:} 
We first drop out the samples that are incorrectly answered by the original Qwen2-VL. Then, we use the performance of the downsampling method as a binary discriminator to categorize the sample into two groups:
\ding{172} Group A: Samples considered as ``difficult'', which are incorrectly answered by the downsampling method.
\ding{173} Group B: Samples considered as ``simple'', which are correctly answered by the downsampling method.
This step effectively tags each sample with the labels of ``simple'' or ``difficult'', filtering the existing benchmarks and removing noisy data that is not applicable for evaluating the visual token compression methods.

\paragraph{Step 3: Result Aggregation:}
Based on the classification in Step 2 and the inference results of visual token compression methods in Step 1, we perform a statistical analysis on the accuracy of the ``difficult'' samples in the methods to be evaluated.
Thus, an indicator that can truly reflect the visual compression methods fairly can be obtained.

In summary, we develop VTC-Bench, a simple but effective framework for evaluating visual token compression methods. The building pipeline of VTC-Bench is shown in Figure~\ref{figuremain}.
Importantly, the VTC-Bench framework can be applied easily to any existing benchmark, transforming it into a more effective benchmark for evaluating visual token compression methods. Meanwhile, the VTC-Bench framework dynamically and reasonably provides a corresponding benchmark subset for each compression ratio, while offering explainable theoretical upper and lower bounds for the final metrics.

\subsection{Evaluation Results \& Discussions}

We conduct extensive experiments across multiple mainstream MLLMs and benchmarks based on VTC-Bench. 
We select Qwen2-VL-7B~\cite{wang2024qwen2} and LLaVA-OV-7B~\cite{li2024llava} as the base MLLMs and evaluate various visual token compression methods (including FastV, VisionZip, PruMerge+, DART) on a subset of ``difficult'' samples filtered by VTC-Bench.
The experimental results are shown in Table \ref{table3} and \ref{table4}, followed by several analysis:

% \noindent \textbf{Downsampling is all your need?} 
% The phenomenon "simple image downsampling consistently outperforms many advanced compression methods across multiple widely used benchmarks" may lead researchers to believe that complex visual token compression methods are superfluous, and directly reducing resolution is a wiser choice.
% When we focus on the "difficult samples" (Group A) selected by the VTC-Bench, the phenomenon mentioned above is reversed. 
% VTC-Bench explicitly tells us that the notion that "downsampling leads to excellent visual token compression performance" is an illusion that arises from the original benchmarks being flooded with numerous "simple samples" that do not require fine-grained details. 
% For the "difficult samples" that truly test a model's visual understanding capabilities, advanced compression methods are not only effective but also necessary.
\noindent \textbf{Is downsampling all you need?}
Across many benchmarks, simple image downsampling often beats more advanced compression methods, suggesting that sophisticated approaches are unnecessary. VTC-Bench overturns this impression: when we restrict evaluation to the compression-relevant ``difficult'' samples (Group~A), the trend reverses. The apparent superiority of downsampling largely stems from original benchmarks being saturated with easy cases that do not require fine-grained cues. By filtering out such samples, VTC-Bench reveals that for truly challenging instances that test visual understanding, advanced compression methods are not only effective but necessary.

\noindent \textbf{What makes an effective benchmark?}
Simple cross-benchmark comparisons (e.g., ``Benchmark~A outperforms Benchmark~B'') only imply that one is harder, without revealing which skills drive the difficulty or whether it is relevant to visual token compression. VTC-Bench addresses this by filtering out samples that do not inform compression performance, yielding an analysis set that is explicitly sensitive to token compression. This suggests a design principle for future work: effective benchmarks for visual token compression should deliberately increase the share of compression-relevant hard cases.

\noindent \textbf{Further expand the accuracy gap}: 
% In VTC-Bench, the performance gaps between methods are further amplified and clarified in many cases. For example, at 75\% compression in ChartQA, the accuracy gap between the VisionZip and FastV methods increases from 8.8\% to 16.2\%. Meanwhile, at 96\% compression in GQA, the accuracy gap between the VisionZip and FastV methods increases from 0.3\% to 9.0\%. 
VTC-Bench amplifies and clarifies method differences. At 75\% compression on ChartQA, the VisionZip–FastV gap widens from 8.8\% to 16.2\%; at 96\% compression on GQA, it grows from 0.3\% to 9.0\%.
These phenomenon effectively indicates that VTC-Bench indeed eliminates data noise unrelated to the visual token compression task, thereby promoting the fairness and effectiveness of the benchmark in the visual token compression task.

%% file: sections/5_conclusion.tex
\section{Conclusion}

This paper systematically analyzes the task mismatch problem presented in current MLLM benchmarks when evaluating visual token compression methods. 
Based on a surprising and counterintuitive finding: simple image downsampling consistently outperforms many advanced compression methods across multiple widely used benchmarks, we conduct a comprehensive empirical study across several advanced visual token compression methods.
Thus, two crucial findings are concluded based on the empirical study: 
\ding{172} Current benchmarks are noisy for the visual token compression task. 
\ding{173} Downsampling can serve as a data filter to evaluate the difficulty of samples upon the visual token compression task.
Furthermore, we propose VTC-Bench, a new evaluation framework specifically designed to optimize and denoise current existing benchmarks by explicitly distinguishing between “simple” and “difficult” samples through downsampling.
Through this work, we hope to not only advance the field of visual token compression but also inspire more discussions within the community on ``how to properly evaluate efficient MLLMs.''

\section{Limitations}

\noindent \textbf{Dependence on a single base model}: The construction of the sample filtering mechanism is entirely dependent on Qwen2-VL. While chosen for its technical necessity, this reliance limits generalizability, as different models may define ``difficult'' samples slightly differently. In the future, this framework will be extended to more base models that have great dynamic resolution support.

\noindent \textbf{Single criterion for filtering}: We use only downsampling as the criterion for identifying ``difficult'' samples. Although effective, employing a more diverse set of baseline methods for ensemble filtering could yield a more robust definition. We will explore a more robust filtering mechanism for the subsequent version of VTC-Bench in future work.

\noindent \textbf{Lack of a formal theoretical definition}: Our identification of ``difficult'' and `simple'' samples is primarily based on the experiment results in  Table~\ref{table1} and \ref{table2}. A formal theoretical framework and mathematical proof for a more explainable sample filtering mechanism is yet to be established.

\noindent \textbf{Limited coverage of methods}: Due to the rapidly evolving field and adaptation constraints, our experimental comparison doesn't include all visual token compression methods. However, the selected methods are sufficient to support our core claims, and the framework remains open and extensible.

\section*{Acknowledgments}
This work was supported by the National Natural Science Foundation of China (Grant No.62506318); Guangdong Provincial Department of Education Project (Grant No.2024KQNCX028); CAAI-Ant Group Research Fund; Scientific Research Projects for the Higher-educational Institutions (Grant No.2024312096), Education Bureau of Guangzhou Municipality; Guangzhou-HKUST(GZ) Joint Funding Program (Grant No.2025A03J3957), Education Bureau of Guangzhou Municipality; the Shanghai Science and Technology Program (Grant No. 25ZR1402278).

%% file: sections/Append.tex
\clearpage
\appendix
\section{Experiment Details}
In this paper, all the experiments are conducted based on one A800 GPU. For the downsampling method and DART, we apply the official code of DART~\cite{wen2025stop}. As to the downsampling method, we resize the image before it enters the MLLM. As to DART, we control the compression ratio through the parameter "Reduction\_Ratio". For VisionZip, PruMerge+, and FastV, we apply the EffiVLM-Bench~\cite{wang2025effivlm}, which offers a unified toolkit to evaluate efficient MLLM. As to these three methods, we control the compression ratio through the parameter "Budget". Considering this paper focuses on the evaluation, it is not related to hyperparameter search. All results come from a single run. The code environment includes Python=3.10, torch=2.6.0, torchvision=0.21.0, and torchaudio=2.6.0. Bicubic interpolation is used to achieve downsampling. We will release all the results, including the output results of each sample and the accuracy results of each benchmark. 

\section{Benchmark Details}

\textbf{GQA:} GQA~\cite{hudson2019gqa} is a large-scale benchmark for visual reasoning and compositional question answering. Based on a strict distribution control, GQA offers 22M valuable reasoning questions.

\textbf{MMBench:} MMBench~\cite{liu2024mmbench} is a comprehensive benchmark designed to evaluate the capabilities of MLLMs. It includes 3217 multiple-choice questions spanning 20 fine-grained dimensions,  supporting several languages such as Chinese and English.

\textbf{MME:} MME~\cite{yin2024survey} provides a systematic framework for evaluating the perceptual and cognitive abilities of MLLMs. It encompasses 14 subtasks across the domains of visual perception, text understanding, reasoning, and cross-modal alignment. 

\textbf{POPE:} POPE~\cite{li2023evaluating} is a benchmark designed to evaluate object hallucination in MLLMs. The pipeline of POPE measures hallucination under random, popular, and adversarial sampling strategies. 

\textbf{MMStar:} MMStar~\cite{chen2024we} is a vision-dependent benchmark for evaluating the reasoning and perception abilities. It has 1500 samples, covering six core abilities with 18 sub-dimensions.

\textbf{OCRBench:} OCRBench~\cite{liu2024ocrbench} is a comprehensive benchmark for evaluating the OCR capabilities of multimodal large models. The benchmark includes 1000 manually verified samples from 29 datasets.

\textbf{ChartQA:} ChartQA~\cite{masry2022chartqa}  evaluates visual and logical reasoning over real-world charts. It includes 9.6k human-written and 23.1k automatically generated questions across different kinds of charts.

\begin{table*}[p]
\small
\renewcommand{\arraystretch}{1.3}
\centering
\caption{Comparison of advanced token compression methods on Qwen2-VL-7B across five reduction ratios. Values are formatted as: \textbf{Group A} ({\scriptsize Group B}) with difference $\Delta$ below. $\Delta$ refers to \textbf{\textit{the absolute gap between groups.}}} 
\label{table_qwen_full}
\renewcommand{\tabcolsep}{4pt}
\resizebox{\textwidth}{!}{
\begin{tabular}{l|c c c c c c c c |c}
\toprule
\textit{\textbf{Method}} & \textit{\textbf{GQA}} & \textit{\textbf{MMB}} & \textit{\textbf{MMB$^{CN}$}} & \textit{\textbf{MME}} & \textit{\textbf{POPE}} & \textit{\textbf{MMStar}} & \textit{\textbf{OCRBench}} & \textit{\textbf{ChartQA}} & \textit{\textbf{Average}} \\
\midrule

% --- 75.00% ---
\rowcolor{lightgrayblue} \multicolumn{10}{c}{\textit{Token Reduction ($\downarrow$ 75.00\%)}} \\ \midrule
\rowcolor{brown!10} + Downsample & \textbf{0.0} {\scriptsize (100.0)} & \textbf{0.0} {\scriptsize (100.0)} & \textbf{0.0} {\scriptsize (100.0)} & \textbf{0.0} {\scriptsize (100.0)} & \textbf{0.0} {\scriptsize (100.0)} & \textbf{0.0} {\scriptsize (100.0)} & \textbf{0.0} {\scriptsize (100.0)} & \textbf{0.0} {\scriptsize (100.0)} & \textbf{0.0} {\scriptsize (100.0)} \\ \midrule
+ FastV~\cite{chen2024image} & \makecell{\textbf{57.8} {\scriptsize (87.6)} \\ \deltayellow{29.8}} & \makecell{\textbf{45.2} {\scriptsize (95.9)} \\ \deltayellow{50.7}} & \makecell{\textbf{56.5} {\scriptsize (95.8)} \\ \deltayellow{39.3}} & \makecell{\textbf{78.9} {\scriptsize (96.7)} \\ \deltayellow{17.8}} & \makecell{\textbf{65.4} {\scriptsize (94.8)} \\ \deltayellow{29.4}} & \makecell{\textbf{41.0} {\scriptsize (76.0)} \\ \deltayellow{35.0}} & \makecell{\textbf{29.1} {\scriptsize (57.2)} \\ \deltayellow{28.1}} & \makecell{\textbf{35.0} {\scriptsize (78.1)} \\ \deltayellow{43.1}} & \makecell{\textbf{51.1} {\scriptsize (85.3)} \\ \deltayellow{34.2}} \\ \cmidrule(lr){2-10}
+ VisionZip~\cite{yang2025visionzip} & \makecell{\textbf{59.3} {\scriptsize (91.2)} \\ \deltayellow{31.9}} & \makecell{\textbf{42.4} {\scriptsize (93.8)} \\ \deltayellow{51.4}} & \makecell{\textbf{42.2} {\scriptsize (93.6)} \\ \deltayellow{51.4}} & \makecell{\textbf{54.9} {\scriptsize (95.3)} \\ \deltayellow{40.4}} & \makecell{\textbf{72.5} {\scriptsize (96.8)} \\ \deltayellow{24.3}} & \makecell{\textbf{45.9} {\scriptsize (81.4)} \\ \deltayellow{35.5}} & \makecell{\textbf{29.6} {\scriptsize (58.1)} \\ \deltayellow{28.5}} & \makecell{\textbf{51.2} {\scriptsize (87.3)} \\ \deltayellow{36.1}} & \makecell{\textbf{49.8} {\scriptsize (87.2)} \\ \deltayellow{37.4}} \\ \cmidrule(lr){2-10}
+ PruMerge+~\cite{shang2024llava} & \makecell{\textbf{57.7} {\scriptsize (91.9)} \\ \deltayellow{34.2}} & \makecell{\textbf{51.2} {\scriptsize (95.1)} \\ \deltayellow{43.9}} & \makecell{\textbf{52.6} {\scriptsize (94.6)} \\ \deltayellow{42.0}} & \makecell{\textbf{62.0} {\scriptsize (95.9)} \\ \deltayellow{33.9}} & \makecell{\textbf{72.1} {\scriptsize (97.5)} \\ \deltayellow{25.4}} & \makecell{\textbf{48.1} {\scriptsize (82.3)} \\ \deltayellow{34.2}} & \makecell{\textbf{21.2} {\scriptsize (46.2)} \\ \deltayellow{25.0}} & \makecell{\textbf{40.5} {\scriptsize (73.6)} \\ \deltayellow{33.1}} & \makecell{\textbf{50.7} {\scriptsize (84.6)} \\ \deltayellow{33.9}} \\ \cmidrule(lr){2-10}
+ DART~\cite{wen2025stop} & \makecell{\textbf{58.9} {\scriptsize (88.1)} \\ \deltayellow{29.2}} & \makecell{\textbf{54.8} {\scriptsize (94.9)} \\ \deltayellow{40.1}} & \makecell{\textbf{52.2} {\scriptsize (94.6)} \\ \deltayellow{42.4}} & \makecell{\textbf{67.6} {\scriptsize (94.9)} \\ \deltayellow{27.3}} & \makecell{\textbf{69.4} {\scriptsize (94.5)} \\ \deltayellow{25.1}} & \makecell{\textbf{47.0} {\scriptsize (77.7)} \\ \deltayellow{30.7}} & \makecell{\textbf{40.2} {\scriptsize (70.2)} \\ \deltayellow{30.0}} & \makecell{\textbf{39.0} {\scriptsize (69.0)} \\ \deltayellow{30.0}} & \makecell{\textbf{53.6} {\scriptsize (85.5)} \\ \deltayellow{31.9}} \\ \midrule

% --- 88.89% ---
\rowcolor{lightgrayblue} \multicolumn{10}{c}{\textit{Token Reduction ($\downarrow$ 88.89\%)}} \\ \midrule
\rowcolor{brown!10} + Downsample & \textbf{0.0} {\scriptsize (100.0)} & \textbf{0.0} {\scriptsize (100.0)} & \textbf{0.0} {\scriptsize (100.0)} & \textbf{0.0} {\scriptsize (100.0)} & \textbf{0.0} {\scriptsize (100.0)} & \textbf{0.0} {\scriptsize (100.0)} & \textbf{0.0} {\scriptsize (100.0)} & \textbf{0.0} {\scriptsize (100.0)} & \textbf{0.0} {\scriptsize (100.0)} \\ \midrule
+ FastV~\cite{chen2024image} & \makecell{\textbf{44.5} {\scriptsize (82.5)} \\ \deltayellow{38.0}} & \makecell{\textbf{39.2} {\scriptsize (90.3)} \\ \deltayellow{51.1}} & \makecell{\textbf{44.1} {\scriptsize (90.8)} \\ \deltayellow{46.7}} & \makecell{\textbf{59.4} {\scriptsize (94.0)} \\ \deltayellow{34.6}} & \makecell{\textbf{46.8} {\scriptsize (88.7)} \\ \deltayellow{41.9}} & \makecell{\textbf{31.0} {\scriptsize (73.0)} \\ \deltayellow{42.0}} & \makecell{\textbf{17.8} {\scriptsize (41.3)} \\ \deltayellow{23.5}} & \makecell{\textbf{28.4} {\scriptsize (61.7)} \\ \deltayellow{33.3}} & \makecell{\textbf{38.9} {\scriptsize (77.8)} \\ \deltayellow{38.9}} \\ \cmidrule(lr){2-10}
+ VisionZip~\cite{yang2025visionzip} & \makecell{\textbf{49.4} {\scriptsize (83.4)} \\ \deltayellow{34.0}} & \makecell{\textbf{33.2} {\scriptsize (89.0)} \\ \deltayellow{55.8}} & \makecell{\textbf{44.4} {\scriptsize (88.1)} \\ \deltayellow{43.7}} & \makecell{\textbf{48.1} {\scriptsize (92.2)} \\ \deltayellow{44.1}} & \makecell{\textbf{70.0} {\scriptsize (92.3)} \\ \deltayellow{22.3}} & \makecell{\textbf{30.3} {\scriptsize (73.0)} \\ \deltayellow{42.7}} & \makecell{\textbf{22.0} {\scriptsize (36.4)} \\ \deltayellow{14.4}} & \makecell{\textbf{49.7} {\scriptsize (74.4)} \\ \deltayellow{24.7}} & \makecell{\textbf{43.4} {\scriptsize (78.6)} \\ \deltayellow{35.2}} \\ \cmidrule(lr){2-10}
+ PruMerge+~\cite{shang2024llava} & \makecell{\textbf{50.4} {\scriptsize (85.8)} \\ \deltayellow{35.4}} & \makecell{\textbf{36.9} {\scriptsize (87.2)} \\ \deltayellow{50.3}} & \makecell{\textbf{38.4} {\scriptsize (86.4)} \\ \deltayellow{48.0}} & \makecell{\textbf{42.9} {\scriptsize (91.9)} \\ \deltayellow{49.0}} & \makecell{\textbf{71.5} {\scriptsize (94.2)} \\ \deltayellow{22.7}} & \makecell{\textbf{28.8} {\scriptsize (71.6)} \\ \deltayellow{42.8}} & \makecell{\textbf{18.1} {\scriptsize (33.0)} \\ \deltayellow{14.9}} & \makecell{\textbf{43.5} {\scriptsize (73.8)} \\ \deltayellow{30.3}} & \makecell{\textbf{41.3} {\scriptsize (78.0)} \\ \deltayellow{36.7}} \\ \cmidrule(lr){2-10}
+ DART~\cite{wen2025stop} & \makecell{\textbf{47.5} {\scriptsize (81.2)} \\ \deltayellow{33.7}} & \makecell{\textbf{40.5} {\scriptsize (87.7)} \\ \deltayellow{47.2}} & \makecell{\textbf{40.9} {\scriptsize (86.9)} \\ \deltayellow{46.0}} & \makecell{\textbf{49.6} {\scriptsize (91.7)} \\ \deltayellow{42.1}} & \makecell{\textbf{57.7} {\scriptsize (90.9)} \\ \deltayellow{33.2}} & \makecell{\textbf{35.4} {\scriptsize (70.0)} \\ \deltayellow{34.6}} & \makecell{\textbf{31.5} {\scriptsize (63.2)} \\ \deltayellow{31.7}} & \makecell{\textbf{27.3} {\scriptsize (57.6)} \\ \deltayellow{30.3}} & \makecell{\textbf{41.3} {\scriptsize (78.6)} \\ \deltayellow{37.3}} \\ \midrule

% --- 93.75% ---
\rowcolor{lightgrayblue} \multicolumn{10}{c}{\textit{Token Reduction ($\downarrow$ 93.75\%)}} \\ \midrule
\rowcolor{brown!10} + Downsample & \textbf{0.0} {\scriptsize (100.0)} & \textbf{0.0} {\scriptsize (100.0)} & \textbf{0.0} {\scriptsize (100.0)} & \textbf{0.0} {\scriptsize (100.0)} & \textbf{0.0} {\scriptsize (100.0)} & \textbf{0.0} {\scriptsize (100.0)} & \textbf{0.0} {\scriptsize (100.0)} & \textbf{0.0} {\scriptsize (100.0)} & \textbf{0.0} {\scriptsize (100.0)} \\ \midrule
+ FastV~\cite{chen2024image} & \makecell{\textbf{35.7} {\scriptsize (81.4)} \\ \deltayellow{45.7}} & \makecell{\textbf{31.9} {\scriptsize (85.7)} \\ \deltayellow{53.8}} & \makecell{\textbf{35.3} {\scriptsize (86.6)} \\ \deltayellow{51.3}} & \makecell{\textbf{48.8} {\scriptsize (91.5)} \\ \deltayellow{42.7}} & \makecell{\textbf{37.4} {\scriptsize (88.1)} \\ \deltayellow{50.7}} & \makecell{\textbf{22.8} {\scriptsize (74.5)} \\ \deltayellow{51.7}} & \makecell{\textbf{13.3} {\scriptsize (33.0)} \\ \deltayellow{19.7}} & \makecell{\textbf{14.8} {\scriptsize (74.8)} \\ \deltayellow{60.0}} & \makecell{\textbf{30.0} {\scriptsize (77.0)} \\ \deltayellow{47.0}} \\ \cmidrule(lr){2-10}
+ VisionZip~\cite{yang2025visionzip} & \makecell{\textbf{41.0} {\scriptsize (79.0)} \\ \deltayellow{38.0}} & \makecell{\textbf{34.5} {\scriptsize (81.9)} \\ \deltayellow{47.4}} & \makecell{\textbf{33.3} {\scriptsize (82.2)} \\ \deltayellow{48.9}} & \makecell{\textbf{43.5} {\scriptsize (88.4)} \\ \deltayellow{44.9}} & \makecell{\textbf{66.3} {\scriptsize (89.4)} \\ \deltayellow{23.1}} & \makecell{\textbf{24.3} {\scriptsize (69.8)} \\ \deltayellow{45.5}} & \makecell{\textbf{14.0} {\scriptsize (25.2)} \\ \deltayellow{11.2}} & \makecell{\textbf{26.1} {\scriptsize (71.3)} \\ \deltayellow{45.2}} & \makecell{\textbf{35.4} {\scriptsize (73.4)} \\ \deltayellow{38.0}} \\ \cmidrule(lr){2-10}
+ PruMerge+~\cite{shang2024llava} & \makecell{\textbf{43.0} {\scriptsize (76.7)} \\ \deltayellow{33.7}} & \makecell{\textbf{29.6} {\scriptsize (76.9)} \\ \deltayellow{47.3}} & \makecell{\textbf{34.1} {\scriptsize (76.1)} \\ \deltayellow{42.0}} & \makecell{\textbf{43.0} {\scriptsize (87.8)} \\ \deltayellow{44.8}} & \makecell{\textbf{67.7} {\scriptsize (87.6)} \\ \deltayellow{19.9}} & \makecell{\textbf{25.5} {\scriptsize (65.5)} \\ \deltayellow{40.0}} & \makecell{\textbf{12.6} {\scriptsize (21.8)} \\ \deltayellow{9.2}} & \makecell{\textbf{29.4} {\scriptsize (68.9)} \\ \deltayellow{39.5}} & \makecell{\textbf{35.6} {\scriptsize (70.2)} \\ \deltayellow{34.6}} \\ \cmidrule(lr){2-10}
+ DART~\cite{wen2025stop} & \makecell{\textbf{41.9} {\scriptsize (78.8)} \\ \deltayellow{36.9}} & \makecell{\textbf{33.8} {\scriptsize (81.8)} \\ \deltayellow{48.0}} & \makecell{\textbf{38.4} {\scriptsize (80.4)} \\ \deltayellow{42.0}} & \makecell{\textbf{46.9} {\scriptsize (88.9)} \\ \deltayellow{42.0}} & \makecell{\textbf{57.0} {\scriptsize (88.5)} \\ \deltayellow{31.5}} & \makecell{\textbf{26.2} {\scriptsize (61.8)} \\ \deltayellow{35.6}} & \makecell{\textbf{25.6} {\scriptsize (57.4)} \\ \deltayellow{31.8}} & \makecell{\textbf{14.5} {\scriptsize (67.1)} \\ \deltayellow{52.6}} & \makecell{\textbf{35.5} {\scriptsize (75.6)} \\ \deltayellow{40.1}} \\ \midrule

% --- 96.00% ---
\rowcolor{lightgrayblue} \multicolumn{10}{c}{\textit{Token Reduction ($\downarrow$ 96.00\%)}} \\ \midrule
\rowcolor{brown!10} + Downsample & \textbf{0.0} {\scriptsize (100.0)} & \textbf{0.0} {\scriptsize (100.0)} & \textbf{0.0} {\scriptsize (100.0)} & \textbf{0.0} {\scriptsize (100.0)} & \textbf{0.0} {\scriptsize (100.0)} & \textbf{0.0} {\scriptsize (100.0)} & \textbf{0.0} {\scriptsize (100.0)} & \textbf{0.0} {\scriptsize (100.0)} & \textbf{0.0} {\scriptsize (100.0)} \\ \midrule
+ FastV~\cite{chen2024image} & \makecell{\textbf{29.6} {\scriptsize (79.2)} \\ \deltayellow{49.6}} & \makecell{\textbf{24.7} {\scriptsize (74.5)} \\ \deltayellow{49.8}} & \makecell{\textbf{33.1} {\scriptsize (74.4)} \\ \deltayellow{41.3}} & \makecell{\textbf{35.6} {\scriptsize (90.1)} \\ \deltayellow{54.5}} & \makecell{\textbf{35.6} {\scriptsize (85.8)} \\ \deltayellow{50.2}} & \makecell{\textbf{21.5} {\scriptsize (68.6)} \\ \deltayellow{47.1}} & \makecell{\textbf{11.0} {\scriptsize (27.5)} \\ \deltayellow{16.5}} & \makecell{\textbf{9.3} {\scriptsize (75.3)} \\ \deltayellow{66.0}} & \makecell{\textbf{25.0} {\scriptsize (71.9)} \\ \deltayellow{46.9}} \\ \cmidrule(lr){2-10}
+ VisionZip~\cite{yang2025visionzip} & \makecell{\textbf{38.6} {\scriptsize (75.6)} \\ \deltayellow{37.0}} & \makecell{\textbf{31.2} {\scriptsize (80.3)} \\ \deltayellow{49.1}} & \makecell{\textbf{32.6} {\scriptsize (80.2)} \\ \deltayellow{47.6}} & \makecell{\textbf{37.9} {\scriptsize (87.4)} \\ \deltayellow{49.5}} & \makecell{\textbf{60.0} {\scriptsize (86.5)} \\ \deltayellow{26.5}} & \makecell{\textbf{24.5} {\scriptsize (69.6)} \\ \deltayellow{45.1}} & \makecell{\textbf{11.0} {\scriptsize (20.4)} \\ \deltayellow{9.4}} & \makecell{\textbf{15.1} {\scriptsize (69.2)} \\ \deltayellow{54.1}} & \makecell{\textbf{31.4} {\scriptsize (71.2)} \\ \deltayellow{39.8}} \\ \cmidrule(lr){2-10}
+ PruMerge+~\cite{shang2024llava} & \makecell{\textbf{38.8} {\scriptsize (72.5)} \\ \deltayellow{33.7}} & \makecell{\textbf{26.0} {\scriptsize (69.3)} \\ \deltayellow{43.3}} & \makecell{\textbf{29.3} {\scriptsize (69.1)} \\ \deltayellow{39.8}} & \makecell{\textbf{37.0} {\scriptsize (83.9)} \\ \deltayellow{46.9}} & \makecell{\textbf{56.4} {\scriptsize (82.2)} \\ \deltayellow{25.8}} & \makecell{\textbf{22.6} {\scriptsize (61.4)} \\ \deltayellow{38.8}} & \makecell{\textbf{9.2} {\scriptsize (18.4)} \\ \deltayellow{9.2}} & \makecell{\textbf{17.0} {\scriptsize (70.3)} \\ \deltayellow{53.3}} & \makecell{\textbf{29.5} {\scriptsize (65.9)} \\ \deltayellow{36.4}} \\ \cmidrule(lr){2-10}
+ DART~\cite{wen2025stop} & \makecell{\textbf{36.4} {\scriptsize (74.5)} \\ \deltayellow{38.1}} & \makecell{\textbf{33.7} {\scriptsize (77.3)} \\ \deltayellow{43.6}} & \makecell{\textbf{36.4} {\scriptsize (75.1)} \\ \deltayellow{38.7}} & \makecell{\textbf{37.9} {\scriptsize (84.9)} \\ \deltayellow{47.0}} & \makecell{\textbf{53.2} {\scriptsize (84.3)} \\ \deltayellow{31.1}} & \makecell{\textbf{22.3} {\scriptsize (63.5)} \\ \deltayellow{41.2}} & \makecell{\textbf{24.7} {\scriptsize (53.7)} \\ \deltayellow{29.0}} & \makecell{\textbf{10.5} {\scriptsize (71.1)} \\ \deltayellow{60.6}} & \makecell{\textbf{31.9} {\scriptsize (73.1)} \\ \deltayellow{41.2}} \\ \midrule

% --- 99.00% ---
\rowcolor{lightgrayblue} \multicolumn{10}{c}{\textit{Token Reduction ($\downarrow$ 99.00\%)}} \\ \midrule
\rowcolor{brown!10} + Downsample & \textbf{0.0} {\scriptsize (100.0)} & \textbf{0.0} {\scriptsize (100.0)} & \textbf{0.0} {\scriptsize (100.0)} & \textbf{0.0} {\scriptsize (100.0)} & \textbf{0.0} {\scriptsize (100.0)} & \textbf{0.0} {\scriptsize (100.0)} & \textbf{0.0} {\scriptsize (100.0)} & \textbf{0.0} {\scriptsize (100.0)} & \textbf{0.0} {\scriptsize (100.0)} \\ \midrule
+ FastV~\cite{chen2024image} & \makecell{\textbf{18.3} {\scriptsize (75.5)} \\ \deltayellow{57.2}} & \makecell{\textbf{18.3} {\scriptsize (55.6)} \\ \deltayellow{37.3}} & \makecell{\textbf{21.5} {\scriptsize (53.2)} \\ \deltayellow{31.7}} & \makecell{\textbf{21.5} {\scriptsize (75.3)} \\ \deltayellow{53.8}} & \makecell{\textbf{44.3} {\scriptsize (59.3)} \\ \deltayellow{15.0}} & \makecell{\textbf{15.0} {\scriptsize (61.0)} \\ \deltayellow{46.0}} & \makecell{\textbf{4.2} {\scriptsize (19.5)} \\ \deltayellow{15.3}} & \makecell{\textbf{3.8} {\scriptsize (73.3)} \\ \deltayellow{69.5}} & \makecell{\textbf{18.4} {\scriptsize (59.1)} \\ \deltayellow{40.7}} \\ \cmidrule(lr){2-10}
+ VisionZip~\cite{yang2025visionzip} & \makecell{\textbf{23.4} {\scriptsize (79.4)} \\ \deltayellow{56.0}} & \makecell{\textbf{28.8} {\scriptsize (78.3)} \\ \deltayellow{49.5}} & \makecell{\textbf{32.2} {\scriptsize (76.9)} \\ \deltayellow{44.7}} & \makecell{\textbf{28.5} {\scriptsize (80.5)} \\ \deltayellow{52.0}} & \makecell{\textbf{53.6} {\scriptsize (70.2)} \\ \deltayellow{16.6}} & \makecell{\textbf{19.4} {\scriptsize (69.4)} \\ \deltayellow{50.0}} & \makecell{\textbf{3.7} {\scriptsize (17.1)} \\ \deltayellow{13.4}} & \makecell{\textbf{5.5} {\scriptsize (70.3)} \\ \deltayellow{64.8}} & \makecell{\textbf{24.4} {\scriptsize (67.8)} \\ \deltayellow{43.4}} \\ \cmidrule(lr){2-10}
+ PruMerge+~\cite{shang2024llava} & \makecell{\textbf{20.7} {\scriptsize (73.9)} \\ \deltayellow{53.2}} & \makecell{\textbf{17.8} {\scriptsize (55.9)} \\ \deltayellow{38.1}} & \makecell{\textbf{21.1} {\scriptsize (52.6)} \\ \deltayellow{31.5}} & \makecell{\textbf{22.9} {\scriptsize (73.9)} \\ \deltayellow{51.0}} & \makecell{\textbf{52.6} {\scriptsize (49.7)} \\ \deltayellow{2.9}} & \makecell{\textbf{17.1} {\scriptsize (57.0)} \\ \deltayellow{39.9}} & \makecell{\textbf{2.5} {\scriptsize (13.0)} \\ \deltayellow{10.5}} & \makecell{\textbf{7.1} {\scriptsize (69.5)} \\ \deltayellow{62.4}} & \makecell{\textbf{20.2} {\scriptsize (55.7)} \\ \deltayellow{35.5}} \\ \cmidrule(lr){2-10}
+ DART~\cite{wen2025stop} & \makecell{\textbf{24.5} {\scriptsize (76.1)} \\ \deltayellow{51.6}} & \makecell{\textbf{26.5} {\scriptsize (63.2)} \\ \deltayellow{36.7}} & \makecell{\textbf{28.1} {\scriptsize (59.0)} \\ \deltayellow{30.9}} & \makecell{\textbf{30.6} {\scriptsize (73.2)} \\ \deltayellow{42.6}} & \makecell{\textbf{41.5} {\scriptsize (67.4)} \\ \deltayellow{25.9}} & \makecell{\textbf{19.2} {\scriptsize (63.7)} \\ \deltayellow{44.5}} & \makecell{\textbf{25.6} {\scriptsize (43.1)} \\ \deltayellow{17.5}} & \makecell{\textbf{4.2} {\scriptsize (70.7)} \\ \deltayellow{66.5}} & \makecell{\textbf{25.0} {\scriptsize (64.6)} \\ \deltayellow{39.6}} \\ 
\bottomrule
\end{tabular}}
\end{table*}

\begin{table*}[p]
\small
\renewcommand{\arraystretch}{1.3}
\centering
\caption{Comparison of advanced token compression methods on LLaVA-OV-7B across five reduction ratios. Values are formatted as: \textbf{Group A} ({\scriptsize Group B}) with difference $\Delta$ below. $\Delta$ refers to \textbf{\textit{the absolute gap between groups.}}}
\label{table_llava_full}
\renewcommand{\tabcolsep}{8pt}
\resizebox{\textwidth}{!}{
\begin{tabular}{l|c c c c c | c}
\toprule
\textit{\textbf{Method}} & \textit{\textbf{GQA}} & \textit{\textbf{MMB}} & \textit{\textbf{MMB$^{CN}$}} & \textit{\textbf{POPE}} & \textit{\textbf{MMStar}} & \textit{\textbf{Average}} \\
\midrule

% --- 75.00% ---
\rowcolor{lightgrayblue} \multicolumn{7}{c}{\textit{Token Reduction ($\downarrow$ 75.00\%)}} \\ \midrule
+ FastV~\cite{chen2024image} & \makecell{\textbf{54.3} {\scriptsize (84.0)} \\ \deltayellow{29.7}} & \makecell{\textbf{70.5} {\scriptsize (93.5)} \\ \deltayellow{23.0}} & \makecell{\textbf{69.1} {\scriptsize (94.7)} \\ \deltayellow{25.6}} & \makecell{\textbf{63.8} {\scriptsize (92.2)} \\ \deltayellow{28.4}} & \makecell{\textbf{48.6} {\scriptsize (73.1)} \\ \deltayellow{24.5}} & \makecell{\textbf{61.3} {\scriptsize (87.5)} \\ \deltayellow{26.2}} \\ \cmidrule(lr){2-7}
+ VisionZip~\cite{yang2025visionzip} & \makecell{\textbf{59.0} {\scriptsize (86.2)} \\ \deltayellow{27.2}} & \makecell{\textbf{67.7} {\scriptsize (93.4)} \\ \deltayellow{25.7}} & \makecell{\textbf{71.3} {\scriptsize (94.2)} \\ \deltayellow{22.9}} & \makecell{\textbf{80.8} {\scriptsize (95.6)} \\ \deltayellow{14.8}} & \makecell{\textbf{44.8} {\scriptsize (62.9)} \\ \deltayellow{18.1}} & \makecell{\textbf{64.7} {\scriptsize (86.5)} \\ \deltayellow{21.8}} \\ \cmidrule(lr){2-7}
+ PruMerge+~\cite{shang2024llava} & \makecell{\textbf{60.4} {\scriptsize (87.1)} \\ \deltayellow{26.7}} & \makecell{\textbf{74.2} {\scriptsize (93.8)} \\ \deltayellow{19.6}} & \makecell{\textbf{73.5} {\scriptsize (94.0)} \\ \deltayellow{20.5}} & \makecell{\textbf{75.6} {\scriptsize (96.3)} \\ \deltayellow{20.7}} & \makecell{\textbf{48.6} {\scriptsize (62.1)} \\ \deltayellow{13.5}} & \makecell{\textbf{66.5} {\scriptsize (86.7)} \\ \deltayellow{20.2}} \\ \midrule

% --- 88.89% ---
\rowcolor{lightgrayblue} \multicolumn{7}{c}{\textit{Token Reduction ($\downarrow$ 88.89\%)}} \\ \midrule
+ FastV~\cite{chen2024image} & \makecell{\textbf{45.3} {\scriptsize (76.1)} \\ \deltayellow{30.8}} & \makecell{\textbf{64.6} {\scriptsize (91.7)} \\ \deltayellow{27.1}} & \makecell{\textbf{66.4} {\scriptsize (92.3)} \\ \deltayellow{25.9}} & \makecell{\textbf{39.1} {\scriptsize (85.2)} \\ \deltayellow{46.1}} & \makecell{\textbf{42.4} {\scriptsize (66.5)} \\ \deltayellow{24.1}} & \makecell{\textbf{51.6} {\scriptsize (82.4)} \\ \deltayellow{30.8}} \\ \cmidrule(lr){2-7}
+ VisionZip~\cite{yang2025visionzip} & \makecell{\textbf{56.6} {\scriptsize (82.8)} \\ \deltayellow{26.2}} & \makecell{\textbf{71.9} {\scriptsize (92.5)} \\ \deltayellow{20.6}} & \makecell{\textbf{71.2} {\scriptsize (92.1)} \\ \deltayellow{20.9}} & \makecell{\textbf{69.6} {\scriptsize (93.6)} \\ \deltayellow{24.0}} & \makecell{\textbf{43.5} {\scriptsize (59.1)} \\ \deltayellow{15.6}} & \makecell{\textbf{62.6} {\scriptsize (84.0)} \\ \deltayellow{21.4}} \\ \cmidrule(lr){2-7}
+ PruMerge+~\cite{shang2024llava} & \makecell{\textbf{57.4} {\scriptsize (82.9)} \\ \deltayellow{25.5}} & \makecell{\textbf{68.8} {\scriptsize (92.8)} \\ \deltayellow{24.0}} & \makecell{\textbf{71.5} {\scriptsize (93.2)} \\ \deltayellow{21.7}} & \makecell{\textbf{76.0} {\scriptsize (93.8)} \\ \deltayellow{17.8}} & \makecell{\textbf{45.8} {\scriptsize (54.9)} \\ \deltayellow{9.1}} & \makecell{\textbf{63.9} {\scriptsize (83.5)} \\ \deltayellow{19.6}} \\ \midrule

% --- 93.75% ---
\rowcolor{lightgrayblue} \multicolumn{7}{c}{\textit{Token Reduction ($\downarrow$ 93.75\%)}} \\ \midrule
+ FastV~\cite{chen2024image} & \makecell{\textbf{36.7} {\scriptsize (73.0)} \\ \deltayellow{36.3}} & \makecell{\textbf{51.2} {\scriptsize (85.8)} \\ \deltayellow{34.6}} & \makecell{\textbf{53.3} {\scriptsize (86.5)} \\ \deltayellow{33.2}} & \makecell{\textbf{29.7} {\scriptsize (81.5)} \\ \deltayellow{51.8}} & \makecell{\textbf{32.6} {\scriptsize (64.3)} \\ \deltayellow{31.7}} & \makecell{\textbf{40.7} {\scriptsize (78.2)} \\ \deltayellow{37.5}} \\ \cmidrule(lr){2-7}
+ VisionZip~\cite{yang2025visionzip} & \makecell{\textbf{49.1} {\scriptsize (79.4)} \\ \deltayellow{30.3}} & \makecell{\textbf{64.3} {\scriptsize (91.2)} \\ \deltayellow{26.9}} & \makecell{\textbf{62.4} {\scriptsize (90.9)} \\ \deltayellow{28.5}} & \makecell{\textbf{53.6} {\scriptsize (90.9)} \\ \deltayellow{37.3}} & \makecell{\textbf{36.6} {\scriptsize (54.6)} \\ \deltayellow{18.0}} & \makecell{\textbf{53.2} {\scriptsize (81.4)} \\ \deltayellow{28.2}} \\ \cmidrule(lr){2-7}
+ PruMerge+~\cite{shang2024llava} & \makecell{\textbf{50.2} {\scriptsize (78.7)} \\ \deltayellow{28.5}} & \makecell{\textbf{66.6} {\scriptsize (91.1)} \\ \deltayellow{24.5}} & \makecell{\textbf{65.3} {\scriptsize (91.0)} \\ \deltayellow{25.7}} & \makecell{\textbf{59.9} {\scriptsize (91.0)} \\ \deltayellow{31.1}} & \makecell{\textbf{34.8} {\scriptsize (53.6)} \\ \deltayellow{18.8}} & \makecell{\textbf{55.4} {\scriptsize (81.1)} \\ \deltayellow{25.7}} \\ \midrule

% --- 96.00% ---
\rowcolor{lightgrayblue} \multicolumn{7}{c}{\textit{Token Reduction ($\downarrow$ 96.00\%)}} \\ \midrule
+ FastV~\cite{chen2024image} & \makecell{\textbf{31.4} {\scriptsize (71.9)} \\ \deltayellow{40.5}} & \makecell{\textbf{37.4} {\scriptsize (77.3)} \\ \deltayellow{39.9}} & \makecell{\textbf{43.1} {\scriptsize (77.9)} \\ \deltayellow{34.8}} & \makecell{\textbf{24.5} {\scriptsize (79.0)} \\ \deltayellow{54.5}} & \makecell{\textbf{28.6} {\scriptsize (57.1)} \\ \deltayellow{28.5}} & \makecell{\textbf{33.0} {\scriptsize (72.6)} \\ \deltayellow{39.6}} \\ \cmidrule(lr){2-7}
+ VisionZip~\cite{yang2025visionzip} & \makecell{\textbf{42.6} {\scriptsize (76.3)} \\ \deltayellow{33.7}} & \makecell{\textbf{55.4} {\scriptsize (86.9)} \\ \deltayellow{31.5}} & \makecell{\textbf{56.9} {\scriptsize (86.8)} \\ \deltayellow{29.9}} & \makecell{\textbf{45.4} {\scriptsize (86.6)} \\ \deltayellow{41.2}} & \makecell{\textbf{30.8} {\scriptsize (49.9)} \\ \deltayellow{19.1}} & \makecell{\textbf{46.2} {\scriptsize (77.3)} \\ \deltayellow{31.1}} \\ \cmidrule(lr){2-7}
+ PruMerge+~\cite{shang2024llava} & \makecell{\textbf{42.7} {\scriptsize (74.5)} \\ \deltayellow{31.8}} & \makecell{\textbf{57.8} {\scriptsize (84.1)} \\ \deltayellow{26.3}} & \makecell{\textbf{59.6} {\scriptsize (84.2)} \\ \deltayellow{24.6}} & \makecell{\textbf{49.9} {\scriptsize (85.9)} \\ \deltayellow{36.0}} & \makecell{\textbf{31.3} {\scriptsize (49.7)} \\ \deltayellow{18.4}} & \makecell{\textbf{48.3} {\scriptsize (75.7)} \\ \deltayellow{27.4}} \\ \midrule

% --- 99.00% ---
\rowcolor{lightgrayblue} \multicolumn{7}{c}{\textit{Token Reduction ($\downarrow$ 99.00\%)}} \\ \midrule
+ FastV~\cite{chen2024image} & \makecell{\textbf{25.7} {\scriptsize (63.0)} \\ \deltayellow{37.3}} & \makecell{\textbf{25.8} {\scriptsize (50.6)} \\ \deltayellow{24.8}} & \makecell{\textbf{29.6} {\scriptsize (46.5)} \\ \deltayellow{16.9}} & \makecell{\textbf{39.3} {\scriptsize (59.0)} \\ \deltayellow{19.7}} & \makecell{\textbf{21.9} {\scriptsize (47.4)} \\ \deltayellow{25.5}} & \makecell{\textbf{28.5} {\scriptsize (53.3)} \\ \deltayellow{24.8}} \\ \cmidrule(lr){2-7}
+ VisionZip~\cite{yang2025visionzip} & \makecell{\textbf{28.3} {\scriptsize (64.4)} \\ \deltayellow{36.1}} & \makecell{\textbf{28.1} {\scriptsize (54.5)} \\ \deltayellow{26.4}} & \makecell{\textbf{32.8} {\scriptsize (53.8)} \\ \deltayellow{21.0}} & \makecell{\textbf{42.1} {\scriptsize (61.8)} \\ \deltayellow{19.7}} & \makecell{\textbf{24.7} {\scriptsize (36.0)} \\ \deltayellow{11.3}} & \makecell{\textbf{31.2} {\scriptsize (54.1)} \\ \deltayellow{22.9}} \\ \cmidrule(lr){2-7}
+ PruMerge+~\cite{shang2024llava} & \makecell{\textbf{25.3} {\scriptsize (60.4)} \\ \deltayellow{35.1}} & \makecell{\textbf{25.5} {\scriptsize (46.9)} \\ \deltayellow{21.4}} & \makecell{\textbf{28.5} {\scriptsize (45.5)} \\ \deltayellow{17.0}} & \makecell{\textbf{40.4} {\scriptsize (56.7)} \\ \deltayellow{16.3}} & \makecell{\textbf{25.2} {\scriptsize (32.6)} \\ \deltayellow{7.4}} & \makecell{\textbf{29.0} {\scriptsize (48.4)} \\ \deltayellow{19.4}} \\ 
\bottomrule
\end{tabular}}
\end{table*}
\section{Complete VTC-Bench Results}

Due to the page limitation, we are unable to offer complete results in the experiment sections. Thus, we provide the evaluation results by group of Qwen2-VL-7B and LLaVA-OV-7B on eight benchmarks here, as shown in Table \ref{table_qwen_full} and \ref{table_llava_full}.

\section{Concurrent Similar Research}
\subsection{VisionThink}
VisionThink~\cite{yang2025visionthink} identifies a key observation: downsampling showns suprising effectiveness on most general VQA tasks except for OCR and detail-sensitive benchmarks. Leveraging this finding, VisionThink is proposed as a novel paradigm that builds an effective and smart LVLM. Instead of applying a fixed compression rate, VisionThink starts inference with a low-resolution image and employs a reinforcement learning framework to enable the model itself to dynamically decide whether the visual information is sufficient or if a high-resolution original image is needed. This approach effectively utilizes the observation by optimizing for efficiency on simple samples while preserving necessary detail for complex, OCR-heavy tasks.

Different from VisionThink's perspective, we further investigate this observation in comprehensive settings and analyze it through the benchmark perspective. 
Through extensive experimental results in Table~\ref{table1} and \ref{table2}, firstly, the task mismatch between current benchmarks and the visual token compression task is clarified. Secondly, the potential of downsampling as a data filter is proven, motivating the subsequent design of VTC-Bench.
\subsection{EffiVLM-Bench}
EffiVLM-Bench~\cite{wang2025effivlm} is a comprehensive evaluation framework for systematically assessing training-free acceleration techniques in LVLMs, including KV cache compression and token prune methods.  
Thanks to the open-source of EffiVLM-Bench, researchers are capable of comparing and testing different compression methods based on the same model without the effort of adapting these methods to different models.

Different from EffiVLM-Bench, we do not aim to build an open-source toolkit for the community, but to reflect whether the benchmarking system for the visual token compression methods. Under such expectations, VTC-Bench is proposed to address the task mismatch between current benchmarks and the visual token compression task.

\begin{figure*}
    \centering
    \includegraphics[width=1\linewidth]{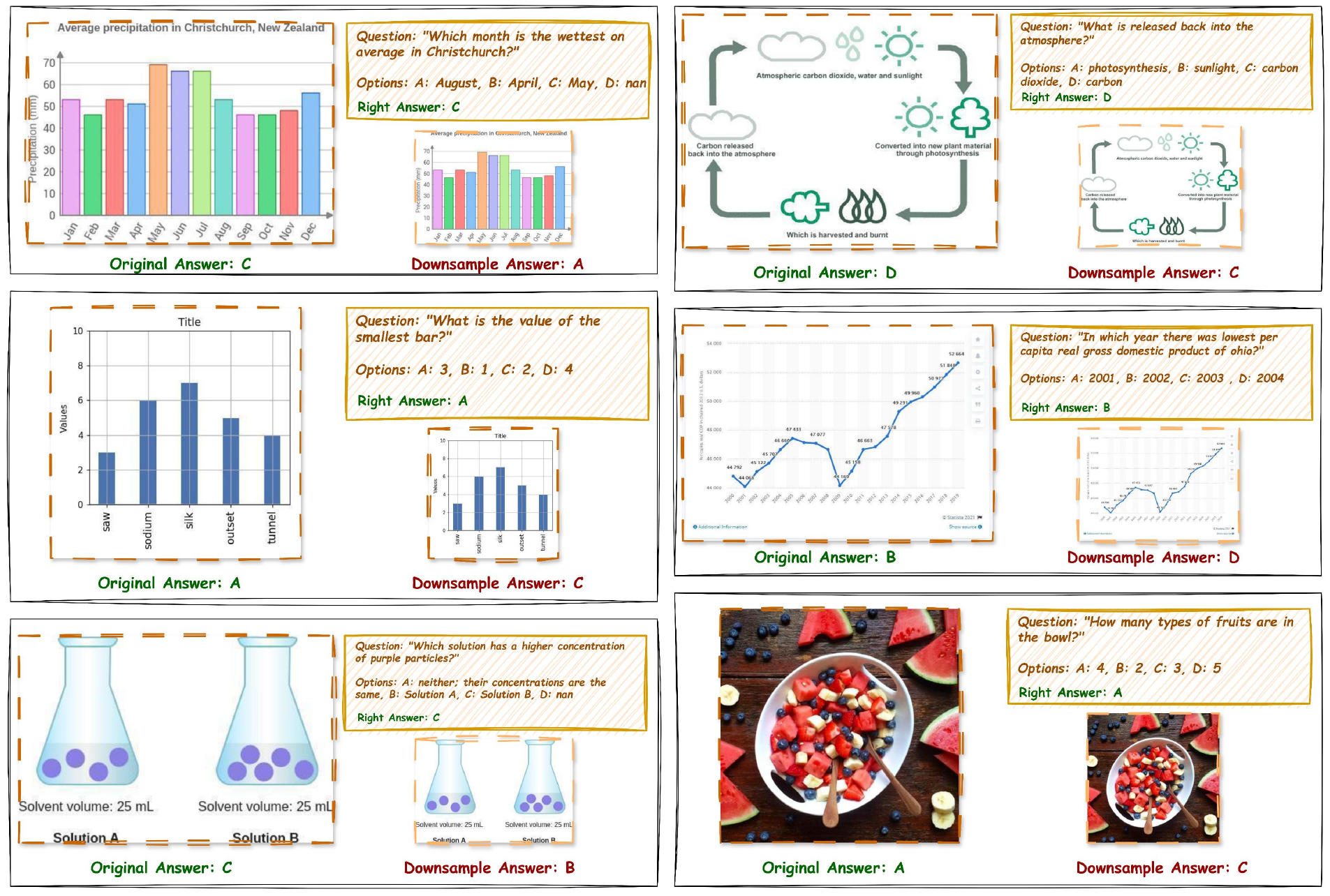}
    \caption{Visualization of ``difficult'' samples with downsampling ratio set to 2.}
    \label{fig:di}
\end{figure*}

\begin{figure*}
    \centering
    \includegraphics[width=1\linewidth]{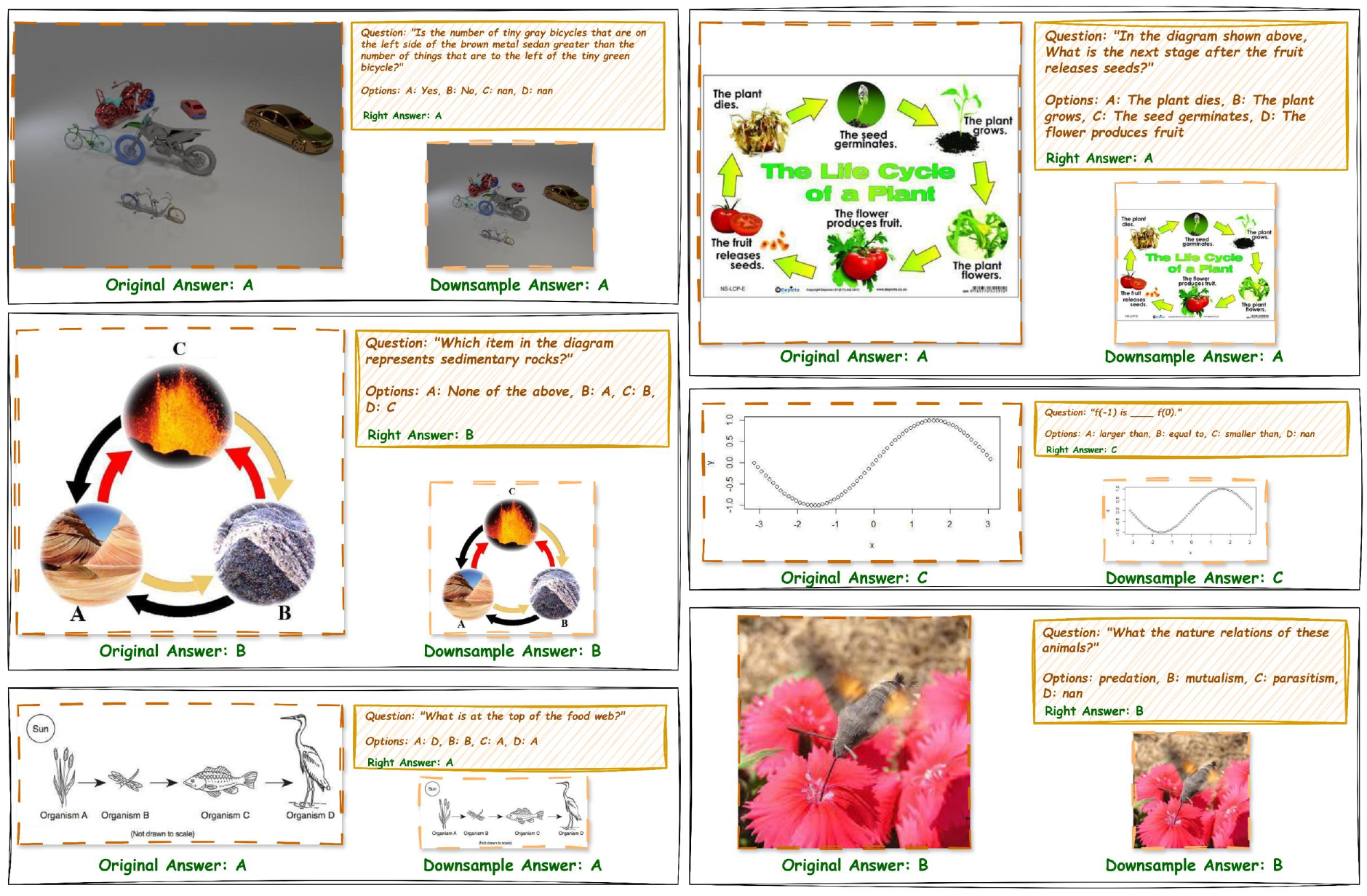}
    \caption{Visualization of ``simple'' samples with downsampling ratio set to 2.}
    \label{fig:si}
\end{figure*}

\section{Visualization between Groups}
Additionally, we visualize the ``difficult'' and ``simple'' samples to observe their characteristics. 
As shown in Figure~\ref{fig:di} and Figure~\ref{fig:si}, the ``difficult'' samples tend to require multiple detail perceptions and comparisons, such as figuring out the extreme values in a chart or performing complex counting. 
For example, Question (Which month is the wettest on average in Christchurch?) requires models to capture several details, including the precipitation of all 12 months, and conduct complex comparisons.
Question (How many types of fruits are in the bowl?) requires models to understand large amounts of tiny fruits, which is a typical complex counting question.
Meanwhile, the ``simple'' samples usually perceive based on medium/large-sized patterns or perform simple dual-value comparisons.
For example, Question (What is at the top of the food web?) only requires understanding four medium-sized patterns. Question (f(-1) is \_ f(0).) requires only the comparison of two values.
In summary, ``difficult'' samples often require answers based on several details, while ``simple'' samples usually require answers based on non-detailed or limited details.

\begin{figure*}[h]
    \centering
    \includegraphics[width=0.9\linewidth]{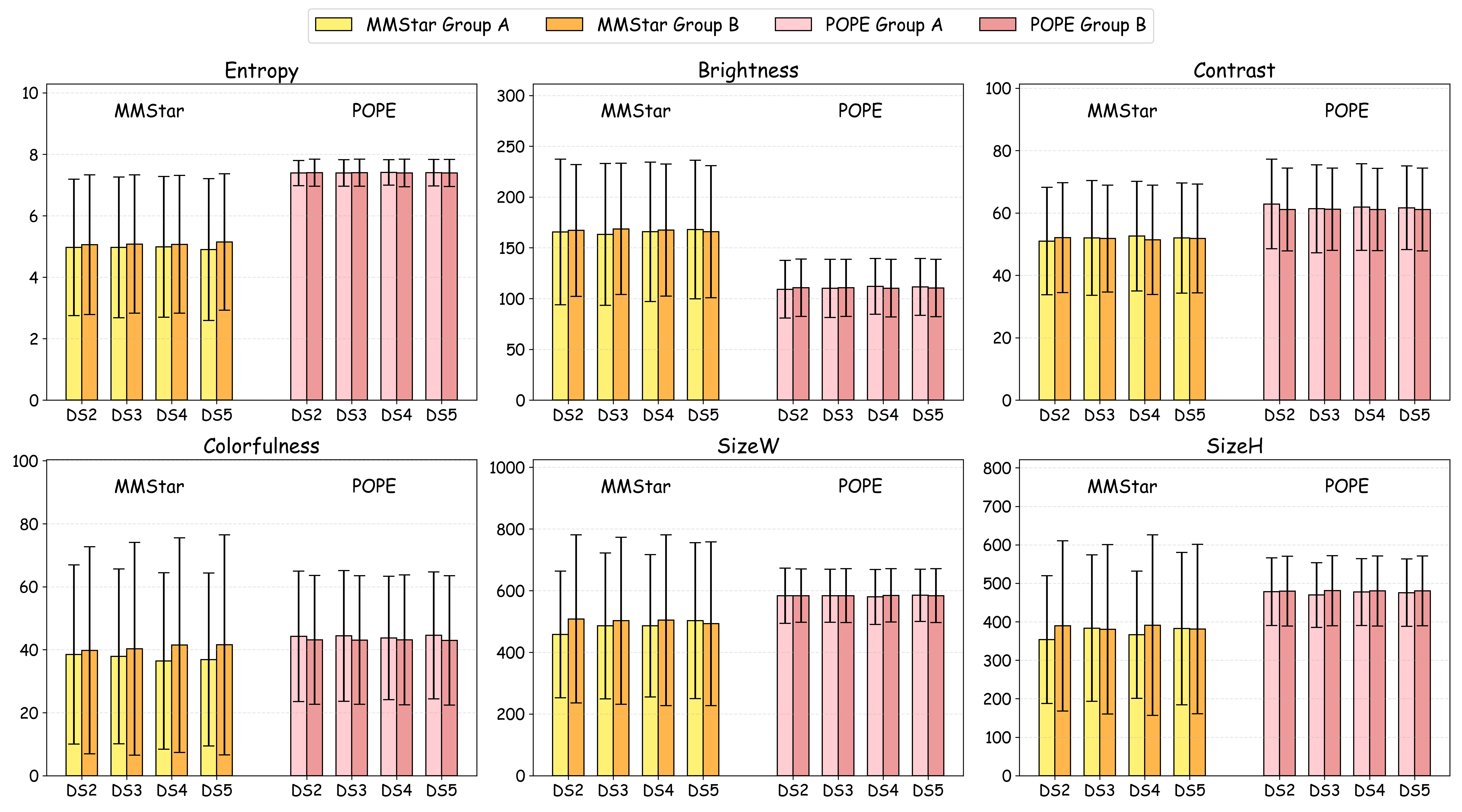}
    \caption{Visualization of low-level visual properties of two groups based on MMStar and POPE across different downsampling ratios. DSx refers to that downsampling ratio is set to x.}
    \label{fig:ana}
\end{figure*}

\begin{table*}[!htbp]
\centering
\caption{Low-level visual properties of two groups based on MMStar and POPE across different downsampling ratios.}
\label{tab:image_analysis}

% 优化视觉效果：增加行高并消除横线间的白缝
\renewcommand{\arraystretch}{1.5}
\setlength{\tabcolsep}{4pt}
\setlength{\aboverulesep}{0pt}
\setlength{\belowrulesep}{0pt}

\resizebox{\textwidth}{!}{%
\begin{tabular}{cccccccc}
\toprule
\textbf{\textit{Benchmark}} & \textbf{\textit{Downsample}} & \textbf{\textit{Group}} & \textbf{\textit{Entropy}} & \textbf{\textit{Brightness}} & \textbf{\textit{Contrast}} & \textbf{\textit{Colorfulness}} & \textbf{\textit{Size ($W \pm \sigma \times H \pm \sigma$)}} \\ 
\midrule

% --- MMStar Section ---
\multirow{8}{*}{\rotatebox{90}{\textbf{MMStar~\cite{chen2024we}}}} 
 & \multirow{2}{*}{Ratio 2} & \gr Group A & \gr $4.97 \pm 2.22$ & \gr $165.63 \pm 71.60$ & \gr $50.99 \pm 17.21$ & \gr $38.51 \pm 28.43$ & \gr $(457.87 \pm 205.37) \times (353.56 \pm 165.86)$ \\
 &                         & Group B & $5.06 \pm 2.27$ & $167.15 \pm 64.92$ & $52.11 \pm 17.62$ & $39.79 \pm 32.87$ & $(508.06 \pm 272.49) \times (389.30 \pm 221.20)$ \\
\cmidrule(lr){2-8}
 & \multirow{2}{*}{Ratio 3} & \gr Group A & \gr $4.97 \pm 2.29$ & \gr $163.13 \pm 69.82$ & \gr $52.00 \pm 18.38$ & \gr $37.91 \pm 27.77$ & \gr $(485.88 \pm 236.38) \times (383.46 \pm 190.36)$ \\
 &                         & Group B & $5.08 \pm 2.25$ & $168.58 \pm 64.67$ & $51.80 \pm 17.12$ & $40.29 \pm 33.75$ & $(502.46 \pm 270.65) \times (380.56 \pm 220.04)$ \\
\cmidrule(lr){2-8}
 & \multirow{2}{*}{Ratio 4} &\gr Group A & \gr $4.99 \pm 2.29$ & \gr $165.73 \pm 68.70$ & \gr $52.57 \pm 17.54$ & \gr $36.42 \pm 28.05$ & \gr $(486.04 \pm 230.52) \times (366.15 \pm 165.08)$ \\
 &                         & Group B & $5.07 \pm 2.24$ & $167.50 \pm 64.95$ & $51.42 \pm 17.52$ & $41.48 \pm 34.06$ & $(504.11 \pm 277.10) \times (391.21 \pm 234.83)$ \\
\cmidrule(lr){2-8}
 & \multirow{2}{*}{Ratio 5} &\gr Group A & \gr $4.90 \pm 2.31$ & \gr $167.97 \pm 68.25$ & \gr $51.97 \pm 17.65$ & \gr $36.88 \pm 27.44$ & \gr $(502.70 \pm 253.08) \times (382.14 \pm 198.24)$ \\
 &                         & Group B & $5.15 \pm 2.22$ & $165.92 \pm 64.96$ & $51.79 \pm 17.45$ & $41.57 \pm 34.94$ & $(492.74 \pm 265.51) \times (380.99 \pm 220.28)$ \\

\midrule \midrule

% --- POPE Section ---
\multirow{8}{*}{\rotatebox{90}{\textbf{POPE~\cite{li2023evaluating}}}} 
 & \multirow{2}{*}{Ratio 2} &\gr Group A & \gr $7.39 \pm 0.41$ & \gr $109.22 \pm 28.33$ & \gr $62.86 \pm 14.36$ & \gr $44.25 \pm 20.71$ & \gr $(583.56 \pm 89.37) \times (478.05 \pm 87.72)$ \\
 &                         & Group B & $7.40 \pm 0.44$ & $110.61 \pm 28.24$ & $61.10 \pm 13.23$ & $43.16 \pm 20.46$ & $(583.88 \pm 86.74) \times (479.65 \pm 90.52)$ \\
\cmidrule(lr){2-8}
 & \multirow{2}{*}{Ratio 3} &\gr Group A & \gr $7.39 \pm 0.43$ & \gr $110.04 \pm 28.62$ & \gr $61.35 \pm 14.10$ & \gr $44.39 \pm 20.77$ & \gr $(583.67 \pm 86.12) \times (469.42 \pm 84.01)$ \\
 &                         & Group B & $7.40 \pm 0.44$ & $110.58 \pm 28.20$ & $61.19 \pm 13.21$ & $43.09 \pm 20.44$ & $(583.89 \pm 86.99) \times (480.72 \pm 90.99)$ \\
\cmidrule(lr){2-8}
 & \multirow{2}{*}{Ratio 4} &\gr Group A & \gr $7.41 \pm 0.41$ & \gr $112.04 \pm 27.46$ & \gr $61.89 \pm 13.85$ & \gr $43.76 \pm 19.59$ & \gr $(579.60 \pm 89.42) \times (477.16 \pm 86.95)$ \\
 &                         & Group B & $7.39 \pm 0.45$ & $110.27 \pm 28.37$ & $61.09 \pm 13.21$ & $43.14 \pm 20.62$ & $(584.58 \pm 86.45) \times (479.95 \pm 90.91)$ \\
\cmidrule(lr){2-8}
 & \multirow{2}{*}{Ratio 5} &\gr Group A & \gr $7.40 \pm 0.43$ & \gr $111.47 \pm 28.01$ & \gr $61.66 \pm 13.43$ & \gr $44.58 \pm 20.15$ & \gr $(584.90 \pm 84.97) \times (475.62 \pm 87.85)$ \\
 &                         & Group B & $7.39 \pm 0.44$ & $110.35 \pm 28.29$ & $61.12 \pm 13.28$ & $42.98 \pm 20.53$ & $(583.67 \pm 87.25) \times (480.28 \pm 90.80)$ \\
\bottomrule
\end{tabular}%
}
\end{table*}
\section{Statistical Analysis of Image Properties between Groups}

To investigate whether the inherent ``difficulty" of a sample for visual token compression could be attributed to basic, low-level visual properties, we conduct a quantitative analysis of key image statistics between samples labeled as ``difficult'' (Group A) and ``simple'' (Group B). The statistics measured included image entropy, brightness, contrast, colorfulness, and original image dimensions. This analysis is performed for both the MMStar and POPE across multiple downsampling ratios used for grouping (from 2 to 5). 

The results in Table~\ref{tab:image_analysis}, and Figure~\ref{fig:ana} reveal a consistent and crucial pattern: there is no statistically significant difference in these low-level image statistics between Group A and Group B for either dataset. 
For instance, in MMStar, the average entropy ranges between 4.90 and 5.15 for both groups, while brightness values are centered around 163-168. Similarly, in POPE, the average entropy remains at approximately 7.40, and the average brightness varies minimally around 110 for both groups. The substantial overlap in the ranges indicated by the standard deviations confirms this lack of separation.

This finding is pivotal. It demonstrates that simple, perceptual image properties cannot explain or easily predict the ``difficulty'' of samples. Samples that are ``difficult'' are not systematically brighter, more colorful, higher in contrast, or larger in size than simpler samples. Consequently, the defining characteristic of a ``difficult'' sample must lie beyond these elementary features.

It is precisely the absence of a straightforward, low-level metric to distinguish difficulty that validates our methodological choice to use downsampling as an operational and effective proxy for identifying samples that challenge the model's visual compression. 
However, although downsampling effectively distinguishes the ``difficulty'' of samples, it is equally important to further find reasonable and high-level metrics to measure difficulty, which requires us to explore further in the future.

\section{Inference Time Comparison}

In this section, we provide a detailed comparison of the inference time between the simple downsampling baseline and one other advanced visual token compression method\footnote{Inference time is collected through our training logs for rough comparison, which is calculated based on the inference start time and end time of all 8 benchmarks.}. 
Considering that different methods optimize code complexity differently and use different acceleration tools, we compare DART and downsampling based on the same code repository from DART within 1 A800. 
As shown in Table~\ref{final}, the reduction in tokens has far less impact on speed than we had anticipated in DART, while downsampling shows more normal results. 
We believe this is most likely due to the time overhead caused by additional metric calculations in the sophisticated visual token compression methods, which further proves the efficiency of downsampling in the previous evaluation system of visual token compression methods.
\begin{table}[h]
\centering
\caption{Inference time comparison of downsampling and DART~\cite{wen2025stop}. \textbf{\textit{Budget}} refers to the proportion of remaining visual tokens.}

% 优化视觉效果：增加行高并消除横线间的白缝

\resizebox{0.23\textwidth}{!}{%
\begin{tabular}{ccc}
\toprule
\textbf{\textit{Budget}} & \textbf{\textit{Downsample}} & \textbf{\textit{DART}} \\
\midrule
1.0000 & 30566s & -\\
0.2500 & 17452s & 25200s\\
0.1111 & 14846s & 24094s\\
0.0625 & 13584s & 24079s\\
0.0400 & 12975s & 23690s\\
0.0100 & 12802s & 23386s\\
\bottomrule
\end{tabular}
}
\label{final}
\end{table}